\documentclass[11pt,a4paper]{article}
\usepackage{times,latexsym}
\usepackage{url}
\usepackage[T1]{fontenc}
\usepackage{float}
\usepackage{microtype}
\usepackage{CJKutf8}

\usepackage{hyperref}
\usepackage{url}

\usepackage{amssymb}
\usepackage{amsmath}
\usepackage{amsthm}
\usepackage{amsfonts}
\usepackage{bm}
\usepackage{nicefrac}
\usepackage{dsfont}
\usepackage{fancyvrb}
\usepackage{fvextra}
\usepackage{listings}
\usepackage{xcolor}
\usepackage{tabularx}% added for table design
\usepackage[all]{nowidow}
\usepackage{xcolor}
\usepackage{colortbl}

\usepackage{booktabs}
\usepackage{multirow}
\usepackage{makecell}
\usepackage{arydshln}

\usepackage{comment}
\usepackage{xspace}
\usepackage{soul}

\usepackage{enumitem}
\usepackage[normalem]{ulem}
\setlist[itemize,enumerate]{leftmargin=*}

\makeatletter
\def\adl@drawiv#1#2#3{%
        \hskip.5\tabcolsep
        \xleaders#3{#2.5\@tempdimb #1{1}#2.5\@tempdimb}%
                #2\z@ plus1fil minus1fil\relax
        \hskip.5\tabcolsep}
\newcommand{\cdashlinelr}[1]{%
  \noalign{\vskip 2pt
           \global\let\@dashdrawstore\adl@draw
           \global\let\adl@draw\adl@drawiv}
  \cdashline{#1}[.4pt/2pt]
  \noalign{\global\let\adl@draw\@dashdrawstore
           \vskip 2pt}}
\makeatother

\definecolor{light-orange}{HTML}{fee9d4}
\definecolor{light-green}{HTML}{d8f0d3}
\definecolor{light-blue}{HTML}{dae8f5}
\definecolor{set10-red}{HTML}{e41a1c}
\definecolor{set10-blue}{HTML}{377eb8}
\definecolor{set10-green}{HTML}{4daf4a}

\definecolor{CustomBlue}{RGB}{57,83,191}
\definecolor{CustomRed}{HTML}{a75151}
\definecolor{BadRed}{HTML}{BF3C54}
\definecolor{GoodGreen}{HTML}{38761D}
\definecolor{DarkGreenOne}{RGB}{106,168,79}
\usepackage[most]{tcolorbox}

\newtcbox{\clustertab}[1]{on line, box align=base, colback={#1},colframe={#1},size=fbox,arc=2pt,top=-1.5pt, bottom=-1.5pt, left=-1.5pt, right=-1.5pt, boxrule=0pt, enlarge left by=1pt}

\newcommand{\firstcluster}{{\footnotesize\clustertab{CustomBlue!60}{1}}}
\newcommand{\secondcluster}{{\footnotesize\clustertab{CustomBlue!40}{2}}}
\newcommand{\thirdcluster}{{\footnotesize\clustertab{CustomBlue!25}{3}}}
\newcommand{\fourthcluster}{{\footnotesize\clustertab{CustomBlue!15}{4}}}
\newcommand{\fifthcluster}
{{\footnotesize\clustertab{CustomBlue!10}{5}}}
\newcommand{\sixthcluster}
{{\footnotesize\clustertab{CustomBlue!8}{6}}}
\newcommand{\seventhcluster}
{{\footnotesize\clustertab{CustomBlue!6}{7}}}
\newcommand{\eighthcluster}
{{\footnotesize\clustertab{CustomBlue!6}{8}}}
\newcommand{\ninthcluster}
{{\footnotesize\clustertab{CustomBlue!6}{9}}}
\newcommand{\tenthcluster}
{{\footnotesize\clustertab{CustomBlue!6}{10}}}
\newcommand{\eleventhcluster}
{{\footnotesize\clustertab{CustomBlue!6}{11}}}
\newcommand{\twelfthcluster}
{{\footnotesize\clustertab{CustomBlue!6}{12}}}
\newcommand{\thirteenthcluster}
{{\footnotesize\clustertab{CustomBlue!6}{13}}}

\newcommand{\tocite}[1]{{\textcolor{orange}{[CITE]}}}

\newcommand{\TowerBase}{\textsc{TowerBase}}
\newcommand{\TowerBlocks}{\textsc{TowerBlocks}}
\newcommand{\TowerChat}{\textsc{TowerChat}}
\newcommand{\TowerInstruct}{\textsc{TowerInstruct}}

\newcommand{\Tower}{\textsc{Tower}}

\newcommand{\comet}{\textsc{Comet}}
\newcommand{\contextcomet}{\textsc{ContextComet}}

\newcommand{\chrf}{\textsc{chrF}}

\newcommand{\mqm}{\textsc{MQM}}

\newcommand{\gptfouro}{\textsc{GPT-4o}}
\newcommand{\gptfour}{\textsc{GPT-4}}

\newcommand{\wmttwofour}{\textsc{WMT24}}
\newcommand{\wmttwothree}{\textsc{WMT23}}
\newcommand{\metricx}{\textsc{MetricX}}

\newcommand{\muda}{\textsc{MuDA}}
\newcommand{\bcontrast}{\textsc{BConTrasT}}

\usepackage{pifont}% http://ctan.org/pkg/pifont
\newcommand{\cmark}{\ding{51}}%
\newcommand{\xmark}{\ding{55}}%

\usepackage[acceptedWithA]{tacl2021v1}
\usepackage{subcaption}
\usepackage{xspace,mfirstuc,tabulary}

\newif\iftaclinstructions
\taclinstructionsfalse % AUTHORS: do NOT set this to true
\iftaclinstructions
\renewcommand{\confidential}{}
\renewcommand{\anonsubtext}{(No author info supplied here, for consistency with
TACL-submission anonymization requirements)}
\newcommand{\instr}
\fi

\iftaclpubformat % this "if" is set by the choice of options

\else

\fi

\title{A Context-aware Framework for Translation-mediated Conversations}

\author{
  José Pombal$^{1,2,3}$\Thanks{Equal contribution.}\,\,, Sweta Agrawal$^{2*}$ 
  \\
  \bf Emmanouil Zaranis$^{2,3}$, Patrick Fernandes$^{2,3,4}$, André F. T. Martins$^{1,2,3,5}$
  \\
  \ \\
  $^1$Unbabel, $^2$Instituto de Telecomunicações
  \\
  $^3$Instituto Superior Técnico, Universidade de lisboa 
  \\
  $^4$Carnegie Mellon University, $^5$ELLIS Unit Lisbon
  \\
  \texttt{jose.pombal@unbabel.com}, \texttt{swetaagrawal20@gmail.com}
}

\date{}

\begin{document}
\maketitle
\begin{abstract}
% Effective communication is fundamental to any interaction, yet challenges arise when participants do not share a common language. 
Automatic translation systems offer a powerful solution to bridge language barriers in 
% such 
scenarios where participants do not share a common language. However, these systems can introduce errors leading to misunderstandings and conversation breakdown. A key issue is that current systems fail to incorporate the rich contextual information necessary to resolve ambiguities and omitted details, resulting in literal, inappropriate, or misaligned translations. In this work, we present a framework to improve large language model-based translation systems by incorporating contextual information in bilingual conversational settings during training and inference. 
% During training, we leverage context-augmented parallel data, which allows the model to generate translations sensitive to conversational history. During inference, we perform quality-aware decoding with context-aware metrics to select the optimal translation from a pool of candidates. 
We validate 
% both components of 
our proposed framework on two task-oriented domains: customer chat and user-assistant interaction. Across both settings, the system produced by our framework---\TowerChat{}---consistently results in better translations than state-of-the-art systems like \gptfouro{} and \TowerInstruct{}, as measured by multiple automatic translation quality metrics on several language pairs. We also show that the resulting model leverages context in an intended and interpretable way, improving consistency between the conveyed message and the generated translations.\footnote{We will release the datasets, outputs and the code to reproduce the findings on acceptance.} 
\end{abstract}

\section{Introduction}

In today’s globalized world, the demand for efficient cross-lingual communication has surged across diverse domains, whether it be for providing global customer support \cite{depalma2006can, zhang-misra-2022-machine}, for enabling real-time multilingual collaboration in meetings \cite{teamworg_gegao, globalmeetings_gegao}, or for facilitating effective patient-doctor interactions \cite{mehandru-etal-2023-physician}.
This need extends beyond human-to-human communication to human-machine interactions, where LLMs have emerged as powerful tools in English but with lacklustre performance in other languages \cite{hu-etal-2023-systematic, jin2024better, etxaniz2023multilingual, etxaniz-etal-2024-multilingual, liu2024translation, dey2024betteraskenglishevaluation}.

One potential solution to bridge this language gap is through translation-mediated conversations. 
In such cases, translation serves as a middle layer between two interacting parties, be it humans or humans and machines. 
In the latter case, for example, instead of relying on the model’s capabilities for addressing user queries in multiple languages (i.e., direct inference), language translation and the downstream task are treated as separate problems (i.e., pretranslation) \cite{etxaniz2023multilingual}.
However, the back-and-forth nature of conversations introduces its own set of challenges, particularly in complex, multi-turn dialogues. 
Context can be lost, cultural nuances overlooked, and translation errors may accumulate over the conversation, leading to misunderstandings or inappropriate responses \cite{tsujii-nagao-1988-dialogue, Robertson2022, mendonca-etal-2023-dialogue}.  

Large language model (LLM)-based translation systems, however, present a promising avenue to address this issue.
Not only are they becoming the state-of-the-art solution for multilingual machine translation (MT)~\cite{zhang2023bayling, wei2023polylm, alves-etal-2023-steering, reinauer2023few-shot, zhu-etal-2024-multilingual, kocmi-etal-2023-findings, kocmi2024preliminary}, but they are also known to handle context adeptly \cite{karpinska-iyyer-2023-large, wang-etal-2023-document-level, he-etal-2024-exploring}. 
Despite their potential, using LLMs to facilitate real-time translation-mediated conversations remains underexplored. 

To tackle this problem, we propose a context-aware framework designed to enhance the translation capabilities of LLMs in conversation settings. 
%
% During training, we use carefully constructed context-augmented examples so that the model can learn to pay more attention to discourse elements like pronoun references, formality, and continuity. We use the original bilingual messages generated by the two interacting parties as context to ensure that the model pays attention to language-specific discourse elements.
During training, we use carefully constructed context-augmented examples, incorporating the original bilingual messages exchanged between the two interacting parties as context. This allows the model to attend to both contextual cues and language-specific discourse elements such as pronoun references, formality, and continuity.
Additionally, we introduce quality-aware decoding \cite[QAD]{fernandes-etal-2022-quality} with context-aware metrics \cite{vernikos-etal-2022-embarrassingly, agrawal2024context} to further help the system prioritize translations that best fit the preceding conversation.

\begin{figure}[t]
    \centering
    \includegraphics[trim={27cm 21cm 27cm 26cm},clip,width=\linewidth]{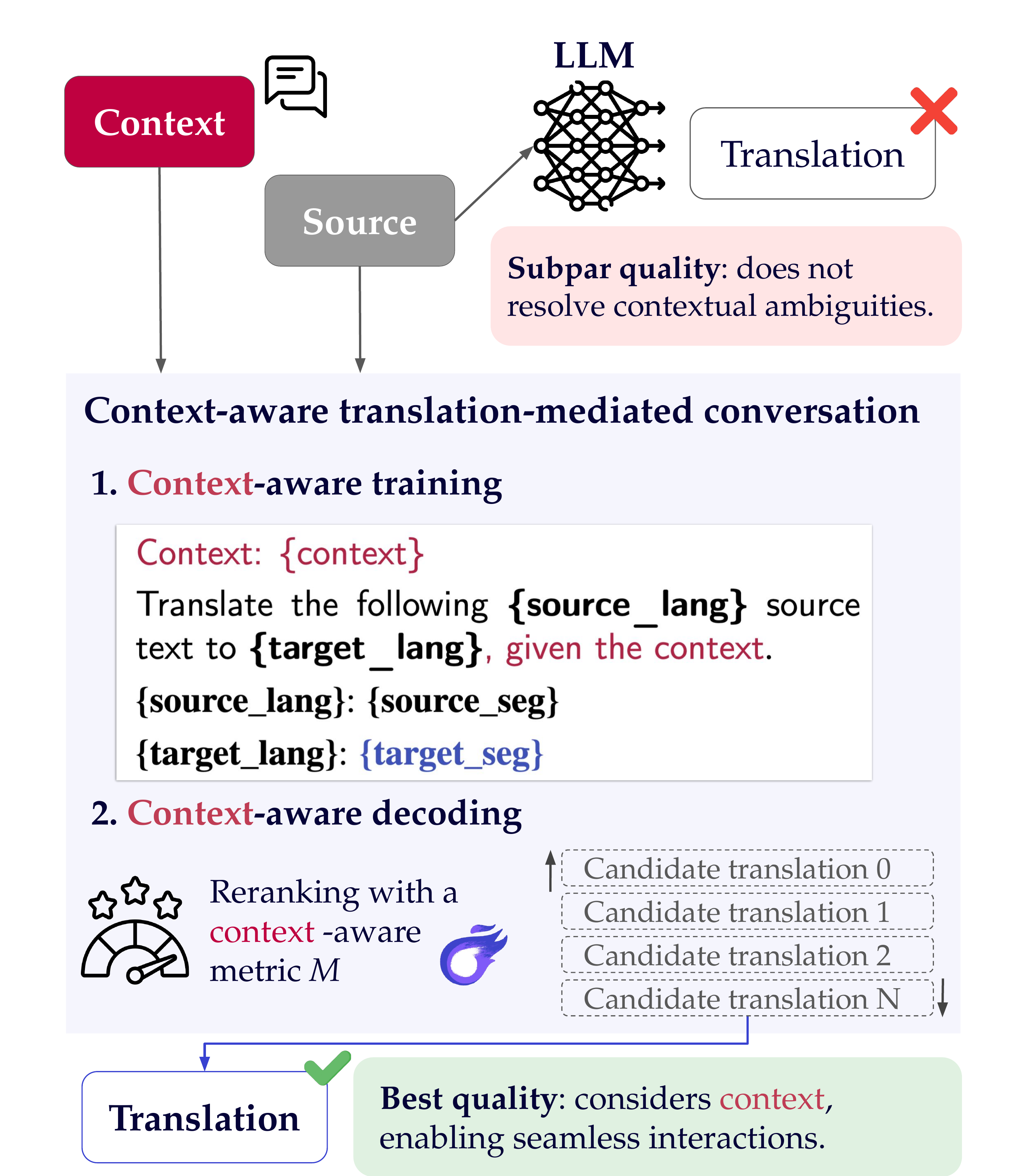}
    \caption{Our framework for optimizing  LLMs for mediating conversations with translation. 
    First, we train the LLM on a conversation translation dataset formatted with context-aware prompts. 
    At inference time, we use a context-aware metric to select the best translation from a pool of candidates. 
    } \label{fig:intro-pic}
\end{figure}

We apply our framework in two bilingual case studies: (1) human-human conversations covering five language pairs: English $\leftrightarrow$ \{German, French, Portuguese, Korean and Dutch\} and (2) human-assistant interactions in English $\leftrightarrow$ German. The assistant functions in its most proficient language, English, while users are supported in their native language via a translation layer. 
Our findings (\S~\ref{sec:results}) reveal that the resulting system, \TowerChat{},  trained using context-augmented instruction training significantly improves translation quality (measured by several automatic metrics), surpassing strong baseline models (\gptfouro{}, \TowerInstruct{}) in multiple language pairs. The gains extend beyond those obtained with in-domain data with standard instructions or with monolingual (English only) context. Additionally, when context is incorporated during inference with quality-aware decoding, our method reduces context-based errors in 9 out of 12 evaluated settings. 

Using existing interpretability tools, we show that our model effectively uses salient parts of the context, particularly in ambiguous sentences, to generate more accurate translations. Finally, our results indicate that context is particularly beneficial when references align well with the surrounding discourse, \textit{i.e.}, they are less surprising given the context. We also observe that the optimal context window varies across language pairs, though incorporating additional turns as context does not degrade performance.
Our contributions are summarized below:
\begin{itemize}
    \item We present a framework that integrates bilingual contextual information in LLM-based translation systems' training and inference stages when translating conversations.
    \item We show the efficacy of our approach in improving the accuracy of translations in multi-turn dialogues, covering human-human and human-assistant interactions.
    \item Our resulting system takes advantage of salient parts of the context to resolve ambiguity, improving the coherence, resolution of pronouns, and contextual accuracy of resulting translations. 
\end{itemize}

\section{Background}

% \sweta{add citations pointed out by reviewer b}

Translation often serves as a vital medium for communication when participants either do not share a common language or opt not to use it. In these situations, context plays a crucial role, directly affecting the quality and appropriateness of translations. Pronoun ambiguity, implicit references, and variations in formality present significant challenges, making accurate translation difficult without contextual cues. Context-aware MT seeks to improve translation quality by considering not just the text itself but the surrounding or broader context, which can involve linguistic, cultural, situational, or even domain-specific information. In MT research, \textit{context} has been interpreted in various ways:  the broader document or neighbouring sentences from which a source to be translated is drawn, the real-world translation setting including the intended audience, the required level of formality, or specialized terminology, among others \cite{Castilho_Knowles_2024}. 

In this work, we define \textit{context} as \textbf{information extending beyond the current turn in bilingual, multi-turn interactions}, crucial for reducing ambiguity and maintaining coherence across turns. Unlike document-level MT, where models can process a complete document at once, translating conversations or dialogues requires turn-by-turn continuity, presenting unique challenges in maintaining consistency across exchanges.\footnote{While document-level MT can also be framed as a form of multi-turn translation, with sentences or blocks of text functioning as individual turns, it fundamentally differs from our setting because it is neither real-time nor bilingual.} We present a review of existing approaches for translating dialogues and how context has been used thus far to improve translation quality in LLMs.

\paragraph{Approaches for Dialogue Translation}  
% \label{ssec:dialogue}
Recognizing the importance of context in dialogues, much prior research has focused on integrating contextual elements or additional meta-information to improve output quality.  For example, \citet{wang-etal-2016-automatic} use speaker tags for modeling the grammatical gender of the participants, while,
\citet{maruf-etal-2018-contextual} incorporate conversation histories into a sentence-based attention model, improving pronoun usage and discourse coherence. 
\citet{liang-etal-2021-modeling} design latent variational modules for learning the distributions of bilingual conversational characteristics (role preference, dialogue coherence, and translation consistency). 
\citet{vincent-etal-2022-controlling-extra} use extra-textual information (the speaker's gender and number of interlocutors) to improve grammatical agreement in dialogue translation. 
While these approaches advance dialogue translation, they rely on task-specific architectural modifications to encoder-decoder models and incorporate only limited aspects of dialogue history. 
In contrast, our proposed framework (\S~\ref{sec:framework}) integrates the full \textit{bilingual} contextual history seamlessly with LLMs to improve translation accuracy in dialogues.

% While these studies highlight ongoing efforts to improve dialogue translation, they also underscore a critical limitation of traditional encoder-decoder models: \textit{the inability to fully and more directly exploit the richness of bilingual conversations}, as standard neural MT models are typically designed for monolingual source-to-target translation, processing sentences in isolation or with limited contextual awareness.
% In contrast, LLMs offer a promising but underexplored alternative. Capable of retaining and utilizing multilingual and crosslingual context across extended text, they are ideal for the context-sensitive nature of bilingual exchanges. Additionally, LLMs' ability to infer implied meanings can enhance conversational fluidity and reduce misunderstandings. To address this gap, we propose a framework that integrates bilingual context during training and inference.

\paragraph{Context-aware MT} \label{ssec:context_aware}
Attempts to include extra-sentential context into the translation process date back to the pre-neural machine translation era~\citep{ws-2013-discourse} and have become increasingly common since~\citep{maruf2021survey, Castilho_Knowles_2024}. However, these efforts have largely focused on cross-sentence context for document-level MT~\citep{wang-etal-2017-exploiting-cross}, where the source text is well-structured and monolingual. Moreover, most models, trained specifically for translation, have shown only marginal improvements over context-agnostic baselines~\citep{lopes-etal-2020-document} and often fail to fully utilize available context~\citep{fernandes-etal-2023-translation}.

% \paragraph{LLMs for Context-aware MT}
Recently, LLMs have shown the potential to effectively use contextual information to perform many NLP tasks, including sentence and document-level translation \cite{karpinska-iyyer-2023-large, wang-etal-2023-document-level}. For instance, \citet{agrawal-etal-2023-context, zhang2023prompting, mu-etal-2023-augmenting} retrieve relevant examples during inference and supply them as context for the current source sentence. Other approaches integrate bilingual dictionaries or domain-specific terminologies \citep{ghazvininejad2023dictionary, moslem-etal-2023-domain} or use prompts to guide LLMs in resolving ambiguity either from the given context \citep{pilault-etal-2023-interactive} or based on pre-existing knowledge~\citep{he-etal-2024-exploring}. Additionally, \citet{treviso2024xtower} propose improving output quality through post-editing of initial drafts with error explanations. \citet{wang-etal-2023-document-level} use context-aware prompts to model document-level translations during inference, whereas, \citet{wu2024adapting} propose training LLMs with document-level context. 

Despite these advances, LLMs' potential to fully exploit bilingual multi-turn contexts remains largely unexplored. Their ability to retain and utilize multilingual and cross-lingual cross-sentence context over extended text makes them well-suited for handling bilingual exchanges.
% Their ability to infer implied meanings can further improve conversational fluidity and reduces errors resulting in miscommunication. 
To address this gap, we propose a framework that integrates bilingual context into LLMs during training and inference in conversational settings.

\section{A Context-aware Framework} \label{sec:framework}

In this section, we outline a framework for effectively leveraging contextual information to improve translation quality in translation-mediated conversations. Our framework addresses both training and inference, showing how context can be systematically integrated to produce translations that align closely with conversational flow.

\subsection{Context-augmented Instruction Finetuning} \label{ssec:context-augment}

\begin{figure}[t]
\begin{center}
\includegraphics[trim={28cm 46cm 28cm 0},clip,width=.9\columnwidth]{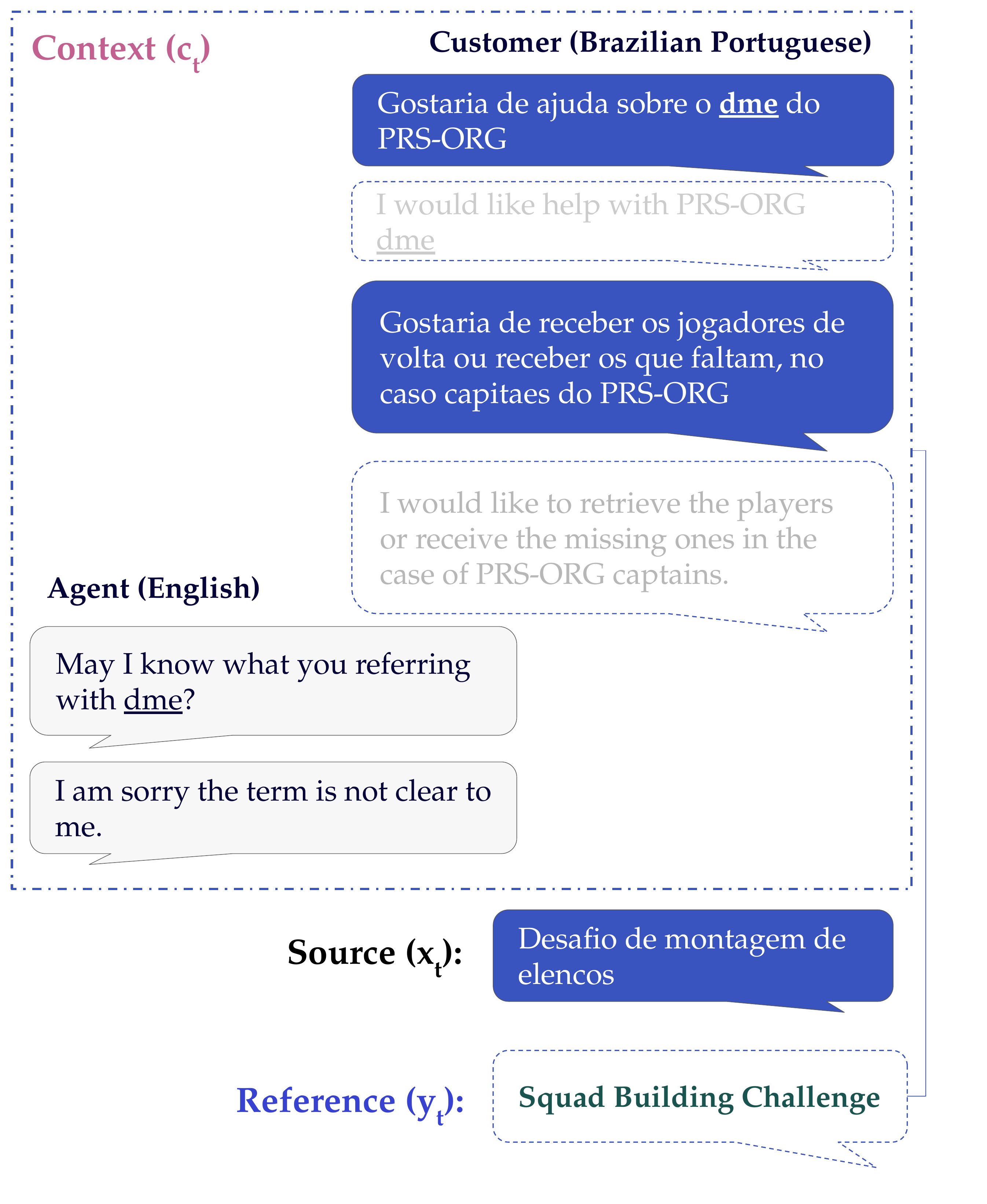}
\end{center}
\caption{A sample conversation illustrating helpfulness of context in resolving ambiguity: correctly translating the source requires inferring that the customer is referring to players (``jogadores'') in previous turns from the ``squad building challenge''. 
% Without context (the turns inside the dotted-blue box), \TowerChat{} mistranslates ``montagem de elencos'' to ``casting''. With context, it correctly translates the source, understanding the customer is talking about a squad building challenge (``dme''), because of the reference to players (``jogadores'') in the customer's second turn.
}
\label{fig:intro-convo}
\end{figure}

Translation of conversations requires understanding both the current utterance and its preceding context. Each instance may be preceded by a series of turns, which, in our case, are bilingual, as shown in Figure~\ref{fig:intro-convo}. To enable the model to effectively leverage context,
we enrich the training dataset with context-augmented instructions. Specifically, for a conversation $C$ of length $L$ with segments $\{(x_t, y_t, c_t)\}_{t=1}^{L}$, where $x_t$ is a text generated by a participant at turn $t$, $y_t$ is its reference translation in the target language, and $c_t$ is the relevant context, we draft a context-augmented instruction as shown in Fig.~\ref{fig:intro-pic}. A training instance is shown in Fig.~\ref{appendix:context-prompt-example}.

\paragraph{Choice of context} The context, 
$c_t$ can be sourced from previous conversational turns, external knowledge bases, or situational factors, encapsulating crucial discourse-level information such as pronoun references and formality. For simplicity, we include only the original \textit{bilingual} texts from the previous turns of the participants, $x_{<t}$. This ensures the model retains a holistic view of the conversation, capturing nuances crucial for accurate translation. However, as LLMs are shown to be more robust when instructed in English~\cite{zhao2024large} and might struggle with understanding the cross-lingual context in low-resource languages, we compare our approach with one where the preceding context is provided solely in English. This requires using generated translations for non-English source texts as context during inference. We hypothesize that while cross-lingual context adds complexity, it better preserves language-specific nuances that might be lost when the context is limited to one language.

Alternative choices can involve extracting or summarizing only the relevant parts of the conversation \cite{krause-etal-2024-graph, sung-etal-2024-context} or incorporating generated translations along with the source texts as a part of the context~\cite{wu2024adapting}. The former adds overhead to the pipeline and risks introducing errors or inconsistencies in the context, potentially losing critical information. Conversely, including system-generated translations during inference creates dependencies on prior outputs, which may lead to errors from inaccuracies that propagate across translations. While our framework provides flexibility in the choice of context, we leave a detailed investigation of alternatives to future work.

\paragraph{Training} We train the model to minimize the cross-entropy loss using a context-aware prompt:
\begin{equation}
    \mathcal{L_\text{ctx}} = - \log P(\textcolor{CustomBlue}{y_t} | x_t, \textcolor{purple}{c_t}).
\end{equation}
This endows the model with the capacity to leverage conversational context when translating.

\subsection{Quality-aware Decoding with Context-aware Metrics} \label{sec:qad}

Decoding strategies informed by translation quality metrics such as Minimum Bayes Risk Decoding (MBR) and Tuned Reranking (TRR) have been shown to improve output quality over greedy decoding~\cite{ stahlberg-etal-2017-neural, fernandes-etal-2022-quality, freitag-etal-2022-high, nowakowski-etal-2022-adam, farinhas-etal-2023-empirical}. In quality-aware decoding, the primary goal is to find a translation among a set of candidates that maximizes an expected utility function, often measured with an automatic MT metric. While previous work has used MBR to improve the quality of individual out-of-context sentences, we extend this by incorporating context into the decoding process. To improve both output quality and contextual accuracy, we apply QAD with context-aware metrics during translation generation, as detailed next.

\paragraph{MBR Decoding} Given a source text, $x_t$, the context, $c_t$, a set of candidate translations sampled from the model, $\mathcal{Y}_t$, and a context-aware metric, $\mathcal{M}$, the utility of each candidate $\hat{y_t} \in \mathcal{Y}_t$, is
\begin{equation}
\small
 u (\hat{y_t}) =  \frac{1}{|\mathcal{Y}|} \sum_{y_t \in \mathcal{Y}_t} \mathcal{M} ( [c_t; x_t], [c_t; y_t], [c_t; \hat{y_t}] ).
\end{equation}
\noindent
To determine the best translation, we then select the candidate that maximizes utility:
\begin{equation}
    y_{\text{mbr}} := \text{arg max}_{\hat{y} \in \mathcal{Y}} [u(\hat{y})].
\end{equation}
Note that the inference strategy can be employed independent of the training, i.e., by sampling from a non-context-aware distribution, $P(y_t|x_t)$ and using a context-aware metric for reranking. This approach can be beneficial when the context-aware metric captures complementary information not fully addressed during training, or when a context-aware MT model is unavailable.

\begin{table*}[]
    \centering
    \renewcommand{\arraystretch}{1.1}
    \footnotesize
    \resizebox{0.98\linewidth}{!}{
    \begin{tabular}{llccccccccccc}
    \toprule
   Dataset & Language Pair & \multicolumn{3}{c}{\# Instances} & \multicolumn{3}{c}{Avg. Source Length}  & \multicolumn{3}{c}{Avg. \# Segments per Conversation}& \multicolumn{2}{c}{\% MuDA tagged}\\
    &  & Train & Dev & Test & Train & Dev & Test& Train & Dev & Test & Dev & Test \\
    \midrule
    \multirow{5}{*}{\wmttwofour{}} & en$\leftrightarrow$de & 17805 & 2569 & 2041 & 47.40 & 52.26 & 53.09 & 36.12 & 31.33 & 30.46 & 15.65 & 15.78\\
   & en$\leftrightarrow$fr & 15027 & 3007 & 2091 & 41.84 & 54.90 & 56.23 & 56.92 & 33.41 & 32.17 & 29.43 & 29.65 \\
    & en$\leftrightarrow$pt-br & 15092 & 2550 & 2040  & 42.72 & 46.46 & 46.49 & 34.69 & 26.56 & 27.95 & 13.02 & 12.99 \\
   & en$\leftrightarrow$ko & 16122 & 1935 & 1982 &39.86 & 47.67 & 46.90 & 38.11 &  50.92 & 47.19 & 0.41 & 0.50\\
   & en$\leftrightarrow$nl & 15463 & 2549 & 2015 & 45.40& 52.31 & 54.31 & 25.99 & 35.40 & 34.74 & 22.01 & 23.13 \\
   \addlinespace[0.3cm]

   \bcontrast & en$\leftrightarrow$de & - & 2100 & -  & - & 43.03 & - & - & 26.92 & -  & - & 22.86 \\
    \bottomrule
    \end{tabular}}
    \caption{Statistics for each language pair and the data splits. 
    % \pombal{we want to make the points: 1) short sentences; 2) some measure of ambiguity (e.g., pronouns); tie with evaluation challenges}
    }
    % \vspace{-0.3cm}
    \label{tab:data_stats}
\end{table*}

\paragraph{Choice of metric $\mathcal{M}$} Standard MT metrics often fall short in effectively utilizing context to determine translation accuracy~\cite{voita-etal-2019-good}. Thus, recent research has focused on designing metrics that better capture discourse information by using inter-sentential context \cite{vernikos-etal-2022-embarrassingly, jiang-etal-2022-blonde, hu-etal-2023-exploring, fernandes-etal-2021-measuring}. For example, context-aware extensions of metrics like \comet{} \cite[\textsc{DocComet} or \contextcomet]{vernikos-etal-2022-embarrassingly, agrawal2024context}, compute quality scores for a source-reference-hypothesis tuple, $(x, y, \hat{y})$, using representations extracted from context-augmented inputs, $([c; x], [c; y], [c; \hat{y}])$, that correlate better with human judgments on document-level MT evaluation. We apply a similar approach by prepending the source from $k$ previous turns, $x_{<t-k:t}$, to the tuple: $([x_{<t-k:t}; x_t], [x_{<t-k:t}; y_t], [x_{<t-k:t}; \hat{y_t}])$. 
% \sweta{add one line clarifying why using hypothesis is not possible for MBR}
Unlike \citet{agrawal2024context}, who use hypotheses in the context, $c_t$, we use the original bilingual context to ensure that the added context remains independent of specific hypotheses from previous turns, enabling stable scoring in MBR decoding by leveraging shared discourse information. 

% \paragraph{Computational Complexity}

\section{Translation-Mediated Conversations: Case Studies}

Translation-mediated conversations have diverse applications across numerous fields, including political, legal, medical, e-commerce and everyday communication. In this work, we focus on two task-oriented applications, as detailed in \S~\ref{ssec:data}. We then present the evaluation setup and experimental settings in \S~\ref{ssec:evaluation} and \S~\ref{ssec:expeirments}, respectively.

\subsection{Application and Datasets} \label{ssec:data}

\paragraph{Customer-support Interaction} 
We use the dataset provided by the WMT 2024 Chat Shared Task~\cite{mohammed-etal-2024-findings}, which includes real bilingual online customer service chats between an English-speaking agent and clients who speak Portuguese, French, Italian, Dutch, or Korean. The dataset spans several domains, including account registration issues, payment and delivery clarifications, and after-sale services in various industries, such as retail and gaming.

\paragraph{Personal-assistant Interaction} We use the \bcontrast{} \cite{farajian-etal-2020-findings} EN-DE dataset based on the Taskmaster-1 \cite{byrne-etal-2019-taskmaster} corpus. The dataset includes task-based bilingual dialogues in six domains: (i) ordering pizza, (ii) creating auto repair appointments, (iii) setting up ride service, (iv) ordering movie tickets, (v) ordering coffee drinks, and (vi) making restaurant reservations. This setup allows us to model structured human-assistant interactions, where the language model facilitates task completion in a controlled yet conversational manner. We note, however, that while we focus on task-oriented dialogue, our framework is designed to generalize to a wide range of LLM-driven interactions, including open-ended dialogue (e.g., with ChatGPT).

The general statistics from both datasets are presented in Table~\ref{tab:data_stats}, including (i) the number of instances in the dataset for each language pair, (ii) the average character length of the source segments, (iii) the average number of segments in a conversation and (iv) the percentage of segments tagged with \muda{}~\cite{fernandes-etal-2023-translation}, an automatic tagger for identifying tokens belonging to certain discourse classes (lexical cohesion, verb forms, pronouns, formality) of potentially ambiguous translations (see Appendix~\ref{app:muda} for more details). Tagging rules are validated by native speakers for linguistic accuracy. For each tag type, we report the F1 score based on matches between tagged words in the reference and hypothesis.
While the \wmttwofour{} development and test sets exhibit a similar distribution regarding segment length and count, they differ significantly from the training dataset.
Furthermore, up to 30\% of en$\leftrightarrow$fr instances are flagged for disambiguation by \muda{}, emphasizing the importance of context for generating high-quality translations.

\subsection{Evaluation} \label{ssec:evaluation}

Ambiguous contextual phenomena that require nuanced evaluation often arise in Chat MT.
As such, we leverage three types of automatic evaluation: 1) for measuring overall translation quality, we use three metrics -- two neural (\comet{}-22 by \citet{rei-etal-2022-comet}, \metricx{}-XL by \citet{juraska-etal-2023-metricx}) and one lexical (\chrf{} by \citet{popovic-2015-chrf});
2) a reference-free LLM-based metric based on \gptfour{} that uses context for providing fine-grained error quality assessment following MQM typology~\cite[ContextMQM]{agrawal2024context};  
3) F1-score on \muda{} tags for measuring whether models correctly resolve lexical ambiguities across diverse discourse phenomena.\footnote{We ignore conversational stopwords when measuring lexical cohesion: \textit{um, uh,  okay,  ok, yes, no}.}

Considering all the metrics is crucial because \comet{} may favour our QAD strategies. 
On Tables~\ref{tab:main_results} and ~\ref{tab:main_results_contrast}, we report performance clusters based on statistically significant performance gaps at 95\% confidence.\footnote{For segment-level metrics, such as \comet{}, we perform significance testing at the segment level. For \chrf{}, we compute corpus-level scores calculated over 100 random samples, each with 50\% of the total segments (without replacement).} We create per-language groups for systems with similar performance, following \citet{freitagetal2023metrics}, and obtain system-level rankings with the average of the obtained clusters, as~\citet{colombo2022best}.
If no model wins on a majority of languages, there is no first cluster.

\subsection{Experimental Settings} \label{ssec:expeirments}

\begin{table*}[t]
    \centering
     \setlength{\tabcolsep}{2pt}
    \renewcommand{\arraystretch}{1.1}
    \footnotesize
    \resizebox{0.98\linewidth}{!}{
    \begin{tabular}{lcrrrcrrr}
    \toprule
    
    && \multicolumn{3}{c}{\textbf{\textsc{en-xx}}} && \multicolumn{3}{c}{\textbf{\textsc{xx-en}}} \\
    
       \cline{3-5}  \cline{7-9} 
    \addlinespace[0.1cm]
    \multirow{-2}{*}{\textbf{\textsc{Model}}} & \multirow{-2}{*}{\textbf{\textsc{Context?}}} & \multicolumn{1}{c}{\textsc{chrF$\uparrow$}} & \multicolumn{1}{c}{\textsc{Comet$\uparrow$}}  & \multicolumn{1}{c}{\textsc{MetricX$\downarrow$}}  && \multicolumn{1}{c}{\textsc{chrF$\uparrow$}} & \multicolumn{1}{c}{\textsc{Comet$\uparrow$}} & \multicolumn{1}{c}{\textsc{MetricX$\downarrow$}} \\
    
    \midrule   
    
   \multicolumn{9}{l}{\small \bf Baselines} \\
    % \nllb{} & \xmark & 59.78 & 88.61 & 1.04  && 70.76 & 88.16 & 0.74 \\   
    \gptfouro & \xmark & 70.09\seventhcluster & 92.62\fifthcluster & 0.37\fifthcluster   && 77.33\fifthcluster & 92.41\fourthcluster & 0.50\thirdcluster  \\
& \cmark  & 70.34\seventhcluster & 92.93\fourthcluster & 0.33\thirdcluster  && 74.75\eighthcluster & 91.59\sixthcluster & 0.58\fourthcluster  \\
\cdashlinelr{1-9}
  \textbf{\TowerInstruct{}} & \xmark & 64.95\tenthcluster & 91.69\seventhcluster & 0.38\fourthcluster   && 76.04\seventhcluster & 92.17\fifthcluster & 0.56\fourthcluster  \\
    & \cmark   & 63.39\twelfthcluster & 91.09\seventhcluster & 0.49\sixthcluster  && 74.32\tenthcluster & 91.36\sixthcluster & 0.60\fourthcluster    \\

     % \TowerInstruct{} 7B  \\ 
    \qquad + QAD (\comet{}) & \xmark  & 65.20\tenthcluster & 92.87\fourthcluster & 0.31\thirdcluster && 75.59\eighthcluster & 92.80\thirdcluster & 0.52\thirdcluster  \\
   \qquad + QAD (\contextcomet{})  & \xmark & 65.06\tenthcluster & 92.57\fifthcluster & 0.31\thirdcluster  && 75.91\seventhcluster & 92.65\fourthcluster & 0.51\thirdcluster  \\

    \cdashlinelr{1-9}
     \textbf{\TowerChat{}} & \xmark & 71.68\seventhcluster & 93.01\fifthcluster & 0.32\thirdcluster &&77.97\fifthcluster & 92.72\fourthcluster & 0.51\thirdcluster  \\
    & \cmark & 75.93\fourthcluster & 93.63\thirdcluster & 0.32\thirdcluster && 78.87\thirdcluster & 93.01\thirdcluster & 0.47\secondcluster  \\
     & \hspace{0.45cm} \cmark (en) & 74.81\fourthcluster & 93.33\thirdcluster & 0.33\thirdcluster && 78.76\thirdcluster & 92.90\fourthcluster & 0.47\secondcluster \\
    \qquad + QAD (\comet{}) & \cmark  & 76.36\secondcluster & \textbf{94.18}\firstcluster & \textbf{0.25}\secondcluster  && \textbf{78.92}\secondcluster & \textbf{93.39}\secondcluster & \textbf{0.44}\firstcluster   \\
    \qquad + QAD (\contextcomet{}) & \cmark  & \textbf{76.56}\secondcluster & 94.05\secondcluster & 0.26\secondcluster && \textbf{78.92}\secondcluster & 93.24\thirdcluster & \textbf{0.44}\firstcluster\\
      \cdashlinelr{1-9}
   \qquad + SFT on QAD (\contextcomet{}) & \xmark & 73.04\sixthcluster & 93.25\fourthcluster & 0.31\thirdcluster && 78.40\fourthcluster & 92.81\fourthcluster & 0.50\secondcluster \\
       & \cmark  & 76.22\thirdcluster & 93.72\thirdcluster & 0.29\thirdcluster && 78.80\thirdcluster & 93.04\thirdcluster & 0.47\secondcluster\\
    % \textbf{\TowerChat{}++} & \xmark & 71.50 & 93.10 & 0.31 && 77.93 & 92.77& 0.51 \\
    %    & \cmark& 75.51 & 93.60 & 0.29 && 78.20 & 93.01& 0.48 \\
    \bottomrule
    \end{tabular}
    }
    \caption{Main Results on WMT24 Chat Shared Task. QAD with \TowerChat{} significantly outperforms all baselines across the board. Models
    are grouped into statistically significant quality clusters. We bold-face both the best overall model and the best \Tower{}-based model for each metric and language pair. 
    }
    \label{tab:main_results}
\end{table*}

\begin{table*}[h]
    \centering
     \setlength{\tabcolsep}{2pt}
    \renewcommand{\arraystretch}{1.2}
    \footnotesize
    \resizebox{0.98\linewidth}{!}{
    \begin{tabular}{lcrrrcrrr}
    \toprule
    &  & \multicolumn{3}{c}{\textbf{\textsc{en-de}}} && \multicolumn{3}{c}{\textbf{\textsc{de-en}}}  \\
    \cline{3-5}  \cline{7-9} 
    \addlinespace[0.1cm]
    \multirow{-2}{*}{\textbf{\textsc{Model}}} & \multirow{-2}{*}{\textbf{\textsc{Context?}}} & \multicolumn{1}{c}{\textsc{chrF$\uparrow$}} & \multicolumn{1}{c}{\textsc{Comet$\uparrow$}}  & \multicolumn{1}{c}{\textsc{MetricX$\downarrow$}}&  & \multicolumn{1}{c}{\textsc{chrF$\uparrow$}} & \multicolumn{1}{c}{\textsc{Comet$\uparrow$}}  & \multicolumn{1}{c}{\textsc{MetricX$\downarrow$}}  \\ 
    \midrule    
     \multicolumn{9}{l}{\small \bf Baselines} \\
     \gptfouro & \xmark & 68.51\secondcluster & 90.60\secondcluster & 0.52\fourthcluster  && 71.14\secondcluster & 92.35\secondcluster & 0.41\firstcluster \\
     & \cmark  & \textbf{70.23}\firstcluster & \textbf{90.96}\firstcluster & \textbf{0.39}\firstcluster&& \textbf{72.72}\firstcluster & \textbf{92.81}\firstcluster & \textbf{0.39}\firstcluster \\
    \cdashlinelr{1-9}
   {\small \bf  \TowerInstruct{}}  & \xmark  & 62.46\thirteenthcluster & 88.20\sixthcluster & 0.54\fourthcluster && 69.54\sixthcluster & 91.98\fourthcluster & 0.47\fourthcluster \\
     & \cmark& 62.75\twelfthcluster & 88.29\sixthcluster & 0.56\fifthcluster && \textbf{70.35}\thirdcluster & 91.99\fourthcluster & 0.46\thirdcluster \\
  \qquad +  QAD (\comet{})    & \xmark  & 63.42\eleventhcluster & 90.15\thirdcluster & 0.46\secondcluster  && 69.24\eighthcluster & 92.31\secondcluster & 0.44\secondcluster \\
   \qquad +  QAD (\contextcomet{})   & \xmark & 64.10\tenthcluster & 89.78\fourthcluster & 0.46\secondcluster  && 69.39\seventhcluster & 92.22\thirdcluster & 0.44\secondcluster \\
    \cdashlinelr{1-9}
     
  {\small \bf  \TowerChat{}}  & \xmark & 65.32\ninthcluster & 89.45\fifthcluster & 0.49\thirdcluster && 68.74\tenthcluster & 91.74\fifthcluster & 0.48\fourthcluster\\
     & \cmark  & 67.09\sixthcluster & 89.60\fourthcluster & 0.46\secondcluster && 69.31\seventhcluster & 92.05\fourthcluster & 0.44\thirdcluster\\
     & \hspace{0.45cm} \cmark (en) & 66.38\seventhcluster & 89.53\fourthcluster & 0.47\thirdcluster && 69.33\seventhcluster & 92.05\fourthcluster& 0.44\secondcluster \\
    \qquad + QAD (\comet{})   & \cmark  & \textbf{67.84}\thirdcluster & \textbf{90.94}\firstcluster & \textbf{0.40}\firstcluster  && 69.71\fourthcluster & \textbf{92.34}\secondcluster & \textbf{0.43}\secondcluster \\
    \qquad +  QAD (\contextcomet{})   & \cmark  & 67.69\fourthcluster & 90.58\secondcluster & \textbf{0.40}\firstcluster && 69.65\fifthcluster & 92.24\thirdcluster & \textbf{0.43}\secondcluster  \\
    \cdashlinelr{1-9}
    
    \qquad + SFT on QAD (\contextcomet{}) & \xmark& 65.92\eighthcluster & 89.70\fourthcluster & 0.46\secondcluster && 69.02\ninthcluster & 91.85\fifthcluster & 0.46\thirdcluster \\
       & \cmark & 67.34\fifthcluster & 89.81\thirdcluster & 0.45\secondcluster && 69.16\eighthcluster & 91.87\fifthcluster & 0.44\secondcluster \\
    % \textbf{\TowerChat{}++} & \xmark  & 64.82 & 89.17 & 0.49 && 68.37 & 91.72& 0.49\\
    %    & \cmark & 66.49 & 89.57 & 0.47 && 68.79 & 91.92& 0.44\\
    \bottomrule
    \end{tabular}}
    \caption{Main Results on \bcontrast. QAD with \TowerChat{} performs comparably with GPT-4o on \comet{} and \metricx{}. Models
    are grouped into statistically significant quality clusters. We bold-face both the best overall model and the best \Tower{}-based model for each metric and language pair. }
    \label{tab:main_results_contrast}
\end{table*}

\paragraph{\TowerChat{}.} We finetune \TowerBase{} 7B with \TowerInstruct{}'s hyperparameters (see Table~\ref{tab:towerchat-hyperparameters}) on the concatenation of \TowerBlocks{} and the training set of the WMT24 shared task using context-aware prompts. 
Importantly, we do not use \bcontrast{} training data. This allows us to assess the model's generalization capabilities to a new domain. 
We report greedy and QAD results with the \TowerChat{}-7B model.
For QAD, we perform MBR with \comet{} or \contextcomet{} on 100 candidates obtained via epsilon sampling with $\epsilon=0.02$~\citep{hewitt-etal-2022-truncation}.
Epsilon sampling restricts token selection to those exceeding a probability threshold $\epsilon$, reducing the risk of generating text that is too unreliable and was shown to be the more effective sampling strategy for MBR over alternatives by \citet{freitag-etal-2023-epsilon}.

\paragraph{Instruction settings.}
To assess whether systems can properly leverage conversational context, we prompt the LLM-based MT with two instruction formats (see Figure~\ref{fig:intro-pic}): 1) \textbf{w/o context} (\xmark), where the model is prompted without any conversational context (without the \textcolor{purple}{purple} highlighted text).  2) \textbf{w/ context} (\cmark), where the  
entire previous bilingual conversation is provided as the context in the prompt.\footnote{During inference, the model generates~\textcolor{CustomBlue}{\textbf{\{target\_seg\}}}.} Additionally, we compare the latter to an alternative where the context is provided only in English: \cmark\textbf{(en)} \---\ retaining original texts in English and using translated texts for non-English source texts, as discussed in Section~\ref{ssec:context-augment}. 
% \sweta{add why use full context here and only 2 for QAD}

\paragraph{Baselines.} We report greedy decoding with  \TowerInstruct{}-7B and \gptfouro.\footnote{We used the snapshot \texttt{gpt-4o-2024-08-06}, with the same prompt as \TowerInstruct{} without a chat template.}
The former serves as a direct baseline for our method, while the latter is a state-of-the-art baseline for MT \cite{sinitsyna-savenkov-2024-comparative}. Furthermore, to assess whether QAD with context-aware metrics can improve translation quality without training, we also report QAD results with \comet{} or \contextcomet{} for \TowerInstruct{}.

\paragraph{Computational Complexity}
MBR scales quadratically with the number of candidates, which can significantly slow down generation. This is particularly relevant in our application, where latency is a concern in real-world deployments. For example, in our setup, with 100 candidates, MBR requires, on average, 100 times more generation steps from the translation model and 10,000 forward passes through the metric. However, the exact computation cost depends on hardware configuration and specific compute optimization. For example, we use the optimized implementation for MBR decoding with \comet{} that uses cached embedding representations.
Recent work has introduced several techniques to improve the efficiency of MBR decoding, including distillation \cite{finkelstein2024mbr}, low-rank factorization \cite{trabelsi2024efficient}, and linear-time approximations \cite{vamvas-sennrich-2024-linear}.
While our primary focus is to show the effectiveness of context-aware QAD in improving translation quality and contextual accuracy, we include an additional experiment that applies training-time self-distillation on MBR outputs~\cite{finkelstein2024mbr}. 
% %
% Specifically, we finetune \TowerChat{} for a second epoch, substituting the original chat data references for the model's own context-aware MBR outputs (SFT on QAD).
% (the last two rows in Tables~\ref{tab:main_results}~and~\ref{tab:main_results_contrast}).
% %
% We show that this single-pass naive distillation improves translation quality by allowing the model to distill MBR benefits without requiring full MBR decoding at runtime, significantly reducing inference-time overhead.

\section{Main Results} \label{sec:results}

Tables~\ref{tab:main_results}~and~\ref{tab:main_results_contrast} present EN$\rightarrow$XX and XX$\rightarrow$EN results on \wmttwofour{} and \bcontrast{}, respectively. 

\paragraph{\TowerChat{} leverages context more adeptly than \TowerInstruct{}.}
One of our goals was to create an LLM-based model that effectively leverages context to generate high-quality translations. 
As shown in Tables~\ref{tab:main_results} and ~\ref{tab:main_results_contrast}, \TowerChat{} outperforms \TowerInstruct{} across both settings (\textit{w/ context} and \textit{w/o context}), language pairs, and evaluation metrics. 
The exception is the \bcontrast{} \textsc{de-en} setting, where \TowerInstruct{} leads in \chrf{}, and lies in the same performance cluster as \TowerChat{} \textit{w/ context}  according to \comet{} and \metricx{}. 
%
% Our empirical analysis suggests that \TowerInstruct{} is especially strong for \textsc{de-en}.
Furthermore, \TowerChat{} shows an average improvement of 4 \chrf{} points for \wmttwofour{} \textsc{en-xx} and 1.7 points \textsc{en-de} when using context (\textit{w/ context}), compared to a context-agnostic prompt (\textit{w/o context}).
This trend also holds when evaluating translation quality with \comet{}, for 8 out of 10 \wmttwofour{} language pairs, as shown in the Appendix Table~\ref{tab:main_results_lps_comet}.
We attribute this to the inclusion of context-augmented instruction dataset in \TowerChat{}'s training, highlighting the effectiveness of in-domain fine-tuning. 
Moreover, incorporating bilingual context consistently improves translation quality over English-only context, yielding an average gain of 1.12 \chrf{} points for WMT24 \textsc{en-xx} settings.

\begin{figure}[t]
    \centering
    \includegraphics[width=0.95\columnwidth, trim={0 4.2cm 0 0},clip]{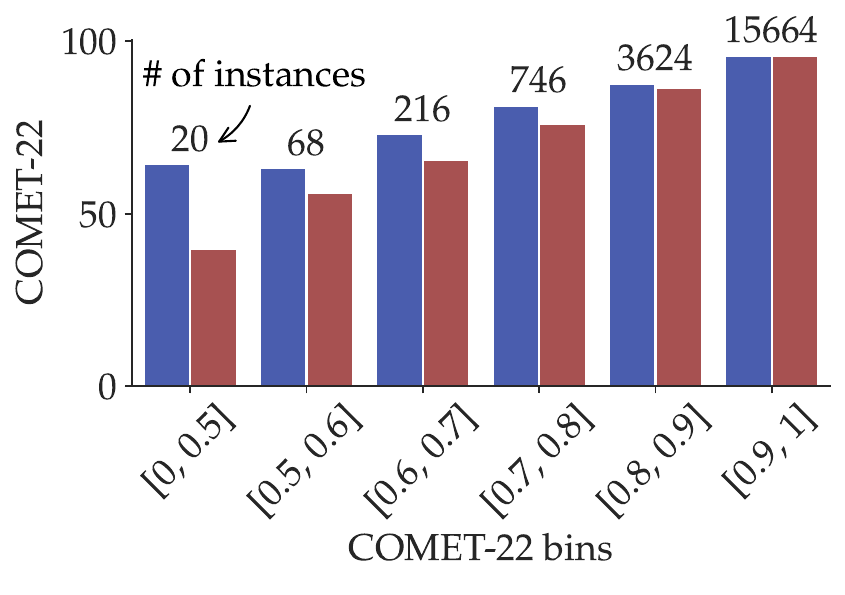}
    \caption{\comet{} from \TowerChat{} \textit{w/o context}. \textcolor{CustomBlue}{Blue}: w/ context. \textcolor{CustomRed}{Red}: w/o context.
    }
    \label{fig:comet-bins-context}
    \vspace{-0.3cm}
\end{figure}

\paragraph{Leveraging contextual information is particularly helpful for low-quality translations.}
\TowerChat{} produces relatively better translations when provided with context than otherwise when the quality of the translation without context is low (see Figure~\ref{fig:comet-bins-context}), decreasing the likelihood of an unsuccessful interaction between participants. Gains are consistent across quality bins on both \textsc{en$\rightarrow$xx} and \textsc{xx$\rightarrow$en} language pairs (see Figure~\ref{fig:comet-bins-context-lps} in Appendix~\ref{appendix-analysis-lps}).  
However, the impact of adding context diminishes as quality increases.
This suggests that by developing quality estimation metrics for segment-level chat translation, one can design a dynamic selection mechanism that applies context only when the estimated quality without context falls below a certain threshold \cite{farinhas2025translate} \---\ we leave this to future work.

\paragraph{QAD results in consistent gains over greedy decoding.} In both datasets and for both Tower models, QAD consistently improves translation quality over greedy decoding across metrics. Furthermore, the highest-quality translations according to all metrics are obtained after performing QAD with \comet{} or \contextcomet{} on top of \TowerChat{}, even outperforming the \gptfouro{}  baseline in the \wmttwofour{} dataset. Notably, the gains extend to \metricx{}, a metric not directly optimized by QAD, highlighting the robustness of the approach. In the \bcontrast{} \textsc{en-de} setting, QAD with \comet{} closes the gap with \metricx{} and \comet{} between \TowerChat{} (greedy) and \gptfouro{} models. This demonstrates how advanced inference methods can improve smaller models, enabling them to compete with larger models like \gptfouro{}.

\begin{table*}[htb!]
    \centering
    \footnotesize
     \setlength{\tabcolsep}{2.5pt}
     \resizebox{\linewidth}{!}{
    \begin{tabular}{llcccccccccccccccccc}
    \toprule
    Model & Context   & \multicolumn{3}{c}{\wmttwofour EN-DE} & \multicolumn{3}{c}{\wmttwofour EN-PT} & \multicolumn{3}{c}{\wmttwofour EN-FR} & \multicolumn{3}{c}{\wmttwofour EN-NL} & \multicolumn{3}{c}{\wmttwofour EN-KO} & \multicolumn{3}{c}{\bcontrast EN-DE} \\
        &    & F1 & XX & EN & F1 & XX & EN& F1& XX & EN& F1 & XX & EN& F1 & XX & EN& F1 & XX & EN \\
    \midrule   
 \TowerInstruct{} & \xmark  & 78.29 & \cellcolor{gray!10}& \cellcolor{gray!10} &76.56 & \cellcolor{gray!10} & \cellcolor{gray!10} &80.44 & \cellcolor{gray!10}& \cellcolor{gray!10}&52.40 & \cellcolor{gray!10} & \cellcolor{gray!10} &52.44 & \cellcolor{gray!10} & \cellcolor{gray!10} &69.47 &\cellcolor{gray!10} & \cellcolor{gray!10}  \\
& \cmark  & 80.12 & \cellcolor{gray!10} & \cellcolor{gray!10} &73.44 & \cellcolor{gray!10} & \cellcolor{gray!10} &80.23 & \cellcolor{gray!10} & \cellcolor{gray!10} &46.46 & \cellcolor{gray!10} & \cellcolor{gray!10} &49.06 & \cellcolor{gray!10} & \cellcolor{gray!10} &69.32 & \cellcolor{gray!10} & \cellcolor{gray!10} \\
\quad + QAD & \xmark &   78.39 &-0.210 &\textbf{-0.458} &\textbf{77.64} &-0.841 & -1.276 &80.47 & \textbf{-0.429} &-0.754 &\textbf{54.80} &-0.384 &-0.709 &54.77 & \textbf{-0.609} & \textbf{-0.862}&\textbf{71.63} & -0.574  & -0.511 \\
\quad + QAD-C & \xmark  &\textbf{78.42} & \textbf{-0.173} &-0.463 &76.47 & \textbf{-0.776} &-1.298 &\textbf{80.62} &-0.455 & \textbf{-0.737} &52.57 & \textbf{-0.375} & \textbf{-0.666} &\textbf{55.62} &-0.666 &-0.870 &69.76 &  \textbf{-0.559}& \textbf{-0.468} \\
\cdashlinelr{1-20}
 \TowerChat{} & \xmark & 77.90 & \cellcolor{gray!10} & \cellcolor{gray!10} &79.53 & \cellcolor{gray!10} & \cellcolor{gray!10} &82.99 & \cellcolor{gray!10} & \cellcolor{gray!10} &49.69 & \cellcolor{gray!10} & \cellcolor{gray!10} &60.53 & \cellcolor{gray!10} & \cellcolor{gray!10} &68.16 & \cellcolor{gray!10} & \cellcolor{gray!10}\\
& \cmark  &79.81 & -0.269 & -0.416 &86.55 & -0.461 & -0.755 &86.34 & -0.410 & -0.635  &68.27 & -0.300 & -0.452 &60.12 & -0.647 & -0.852 &\textbf{76.92} & -0.578 & -0.476 \\
 
% & \cmark  &79.81 & \cellcolor{gray!10} & \cellcolor{gray!10} &86.55 & \cellcolor{gray!10} & \cellcolor{gray!10} &86.34 & \cellcolor{gray!10} & \cellcolor{gray!10} &68.27 & \cellcolor{gray!10} & \cellcolor{gray!10} &60.12 & \cellcolor{gray!10} & \cellcolor{gray!10} &\textbf{76.92} & \cellcolor{gray!10} & \cellcolor{gray!10} \\

\quad + QAD & \cmark   &80.02 &-0.176 &-0.351 &{87.24} &-0.441 &{-0.629} &87.11 & \textbf{-0.267} &-0.664 &76.96 &-0.236 & \textbf{-0.457} &62.50 &-0.349 & \textbf{-0.874} &76.56 & -0.417& -0.416 \\
\quad + QAD-C & \cmark &{80.68} & \textbf{-0.153} &\textbf{-0.331} &85.94 &\textbf{-0.438} & \textbf{-0.600} &{87.17} &-0.313 & \textbf{-0.638} &\textbf{78.74} & \textbf{-0.219} &-0.481 &\textbf{64.34} & \textbf{-0.318} & -0.948  &74.50 & \textbf{-0.397}& \textbf{-0.408}\\
\quad  + SFT (QAD-C) & \cmark & \textbf{84.11}  & -0.260 & -0.407 & \textbf{87.49}&-0.501 & -0.608 &\textbf{87.41} & -0.388 & -0.635 &71.07 & -0.253 & -0.503 &63.94  &  -0.430 & -0.906 &76.12 & -0.591 & -0.461 \\
    \bottomrule
    \end{tabular}
    }
    \caption{Context-based Evaluation. QAD with \contextcomet{} (QAD-C) outperforms QAD with \comet{} (QAD) on MuDA F1 and ContextMQM in 7/12 and 15/24 settings, respectively. 
    }
    \label{tab:context_results}
\end{table*}

\begin{table*}[!t]
    \centering
    \footnotesize
    \resizebox{0.95\linewidth}{!}{
    \begin{tabular}{cccrrrcrrr}
    \toprule
    
    \multicolumn{2}{c}{\textbf{\textsc{Context-aware?}}} && \multicolumn{3}{c}{\textbf{\textsc{en-xx}}} && \multicolumn{3}{c}{\textbf{\textsc{xx-en}}} \\
    
    \cline{1-2} \cline{4-6} \cline{8-10}
    \addlinespace[0.1cm]
    Training & Inference && \multicolumn{1}{c}{\textsc{chrF$\uparrow$}} & \multicolumn{1}{c}{\textsc{Comet$\uparrow$}}  & \multicolumn{1}{c}{\textsc{MetricX$\downarrow$}}  && \multicolumn{1}{c}{\textsc{chrF$\uparrow$}} & \multicolumn{1}{c}{\textsc{Comet$\uparrow$}} & \multicolumn{1}{c}{\textsc{MetricX$\downarrow$}} \\
    \midrule   
    \xmark& \xmark && 75.80 & 93.54 & \textbf{0.31} && 77.80 & 92.80& 0.50  \\
    \xmark& \cmark && 72.89 & 92.73 & 0.38 && 75.39 & 91.21& 0.59  \\
   
    \cdashlinelr{1-10}
    \cmark & \xmark && 71.68 & 93.01 & 0.32 && 77.97 & 92.72 & 0.51 \\
    \cmark& \cmark && \textbf{75.93}& \textbf{93.63} & 0.32 && \textbf{78.87} & \textbf{93.01} & \textbf{0.47} \\

    \bottomrule
    \end{tabular}
    }
    \caption{Ablation of using context-aware prompts during training and/or inference when \TowerInstruct{} is finetuned with TowerBlocks and WMT24 Chat Datasets: \TowerChat{}, trained with context-aware prompts, results in the best overall quality. 
    % \patrick{This table is not self-descriptive atm i.e. its unclear that this is in-domain ablation Maybe rename the first row?}\sweta{Updated it, not sure if its any better}
    }
    \label{tab:context-aware-ablation}
\end{table*}

\paragraph{\muda{}-based evaluation validates our findings.} We report \muda{} F1 scores between references and generated hypotheses for a subset of models in Table~\ref{tab:context_results}.\footnote{Phenomena-specific plots are presented in Figure~\ref{fig:muda-accuracy-lps}.} We can observe that, on average, across phenomena, on all datasets and language pairs: 1) \TowerChat{} \textit{w/ context} achieves higher F1 than \TowerInstruct{} \textit{w/o context}, confirming that \TowerChat{} uses context to improve accuracy on discourse phenomena; 2) QAD-based approaches improve upon their respective greedy decoding counterparts. In 7 out of 12 settings (including QAD with both \TowerChat{} and \TowerInstruct{}), QAD with \contextcomet{} (QAD-C) achieves higher F1 than QAD with \comet{} (QAD). This is also reflected in \chrf{} scores per language pair (Appendix Table~\ref{tab:main_results_lps_chrf}), where QAD-C outperforms QAD in 7 out of 10 settings.

\paragraph{Fine-grained error analysis shows QAD-C performs better with context-aware hypotheses.} 
To compare QAD and QAD-C, we obtain fine-grained MQM-like assessments with ContextMQM referred to as MQM (Table~\ref{tab:context_results}). Contrary to \muda{}, MQM can identify errors that go beyond surface-level properties.  Overall, leveraging QAD-C with \TowerInstruct{} results in similar MQM scores as QAD on average; however, when the pool of candidates includes context-aware hypotheses as with \TowerChat{}, QAD-C outperforms QAD on MQM evaluation in 9 out of 12 settings. To explain the effectiveness of QAD-C with \TowerChat{}, we compare the quality of hypotheses generated by the two models as measured by \chrf{} against references in Appendix~\S~\ref{sec:oracle}: the candidates generated by \TowerChat{} have a higher overlap with the reference than those generated by \TowerInstruct{}.
This shows that leveraging better context-aware metrics should further improve translation quality.

\begin{table}[t]
\centering
\setlength{\tabcolsep}{3pt}
\footnotesize
\begin{tabular}{l c c@{\hspace{.2cm}} c}
\toprule
 % & \multicolumn{3}{c}{\wmttwothree{}} \\
 Models & \multicolumn{1}{c}{\textsc{en$\rightarrow$xx}} & & \multicolumn{1}{c}\textsc{xx$\rightarrow$en} \\
\midrule
\footnotesize{\TowerInstruct{}-7B} & \footnotesize{84.28} & & \footnotesize{82.77} \\
\footnotesize{\TowerChat{}-7B} & \footnotesize{83.95} & & \footnotesize{82.54} \\
\bottomrule
\end{tabular}
% \end{center}
\caption{\comet{} scores for \TowerInstruct{} and \TowerChat{} on the \wmttwothree{} test set.}
\label{tab:general-translation}
\vspace{-0.3cm}
\end{table}

\paragraph{Distilling MBR outputs can improve quality while reducing computational overhead.}
Having established the benefits of QAD with \contextcomet{} in enhancing contextual accuracy—measured by improvements in \muda{} and reductions in errors according to \mqm \---\ we now evaluate the impact of distilling this information via SFT using the WMT24 chat shared task dataset. Specifically, we finetune \TowerChat{} for a second epoch, replacing the original chat data references with the model’s own context-aware MBR outputs (SFT on QAD). This simple, single-pass distillation yields quality improvements over \TowerChat{} across automatic metrics in out-of-English settings (see last two rows in Tables \ref{tab:main_results} and \ref{tab:main_results_contrast}) while also enhancing \muda{} accuracy (Table~\ref{tab:context_results}). These results prove that further efficiency gains in such systems are possible, though we leave deeper exploration of this to future work.

\paragraph{Context-augmented instructions drive improvements beyond in-domain training. }
We conduct an additional ablation to confirm that the observed improvements are driven by the context-augmented instruction structure rather than by the exposure to in-domain data. In this setup, \TowerInstruct{} is fine-tuned on a concatenation of \TowerBlocks{} and the full WMT24 training dataset without any context-aware prompts (referred to as \TowerInstruct{} (Chat)). As shown in Table~\ref{tab:context-aware-ablation}, the context-aware \TowerChat{} model consistently outperforms the non-contextual \TowerInstruct{} (Chat) across both \textsc{en-xx} and \textsc{xx-en} directions, particularly for the latter. This is particularly evident when context-aware prompts are used during inference, where \TowerChat{} achieves notable improvements in \chrf{} and \comet{} scores and maintains competitive \metricx{} values. These results highlight that simply incorporating improved context-aware instructions during training, using the same in-domain dataset, enables the model to better focus on fine-grained details in the output, thereby enhancing its contextual understanding. Hence, it is not merely the quantity of domain-relevant data but how the model is guided to leverage contextual information that drives improvements.

\paragraph{Finetuning on chat data does not degrade general translation capabilities.}
To ensure that training on chat data did not impact translation capabilities on generic data, we report \comet{} on the standard WMT23 benchmark \cite{kocmi-etal-2023-findings} averaged across \textsc{en$\rightarrow$xx} and \textsc{xx$\rightarrow$en} directions for \TowerInstruct{} and \TowerChat{} in 
Table~\ref{tab:general-translation}. \TowerChat{} suffers only minor degradation ($-0.3$) relative to \TowerInstruct{}.

\iffalse
\begin{figure*}[t] 
    \centering
    \includegraphics[width=0.35\linewidth,trim={0 0 7.12cm 0},clip]{figures/turn_scores_CHRF.pdf} 
    % 
    \includegraphics[width=0.45\linewidth]{figures/turn_scores_COMET.pdf} 
    \caption{Comparison of \chrf{} and \comet{} scores for different amounts of turns included in the prompt.}
    %
    \label{fig:turn-analysis}
\end{figure*}
\fi

\section{Assessing Context Usage}\label{sec:assessing}

While our results show that using context in training and inference improves translation quality, it is still not clear when how context influences specific translations.
As incorporating context in the translation process incurs additional computational costs, understanding when the context is used (beneficially) and what parts of this context are most influential could allow a more selective and efficient use of context. Additionally, validating that models use context in interpretable ways builds confidence in their reliability for real-world applications~\cite{yin-neubig-2022-interpreting, briakou-etal-2023-explaining,sarti2024quantifying, cohen-wang2024contextcite}. 
Using the \wmttwofour{} Chat dataset, we analyze how much context is relevant for generating accurate and higher-quality translations, then assess whether the context is used meaningfully by \TowerChat{}. 

\subsection{How much context is needed?}

\begin{figure}[t] 
    \centering
    \includegraphics[width=0.85\linewidth]{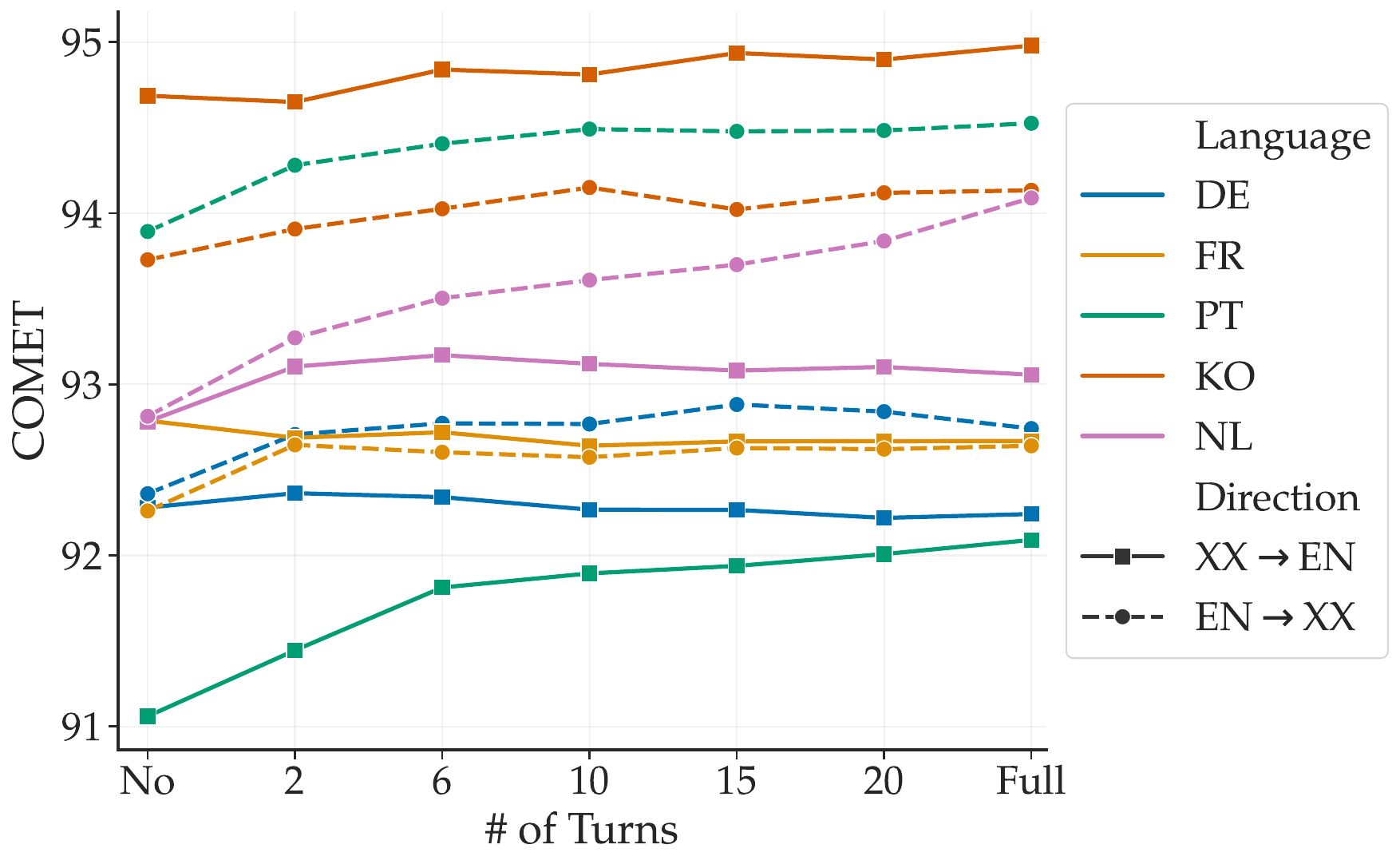} 
     \includegraphics[width=0.85\linewidth]{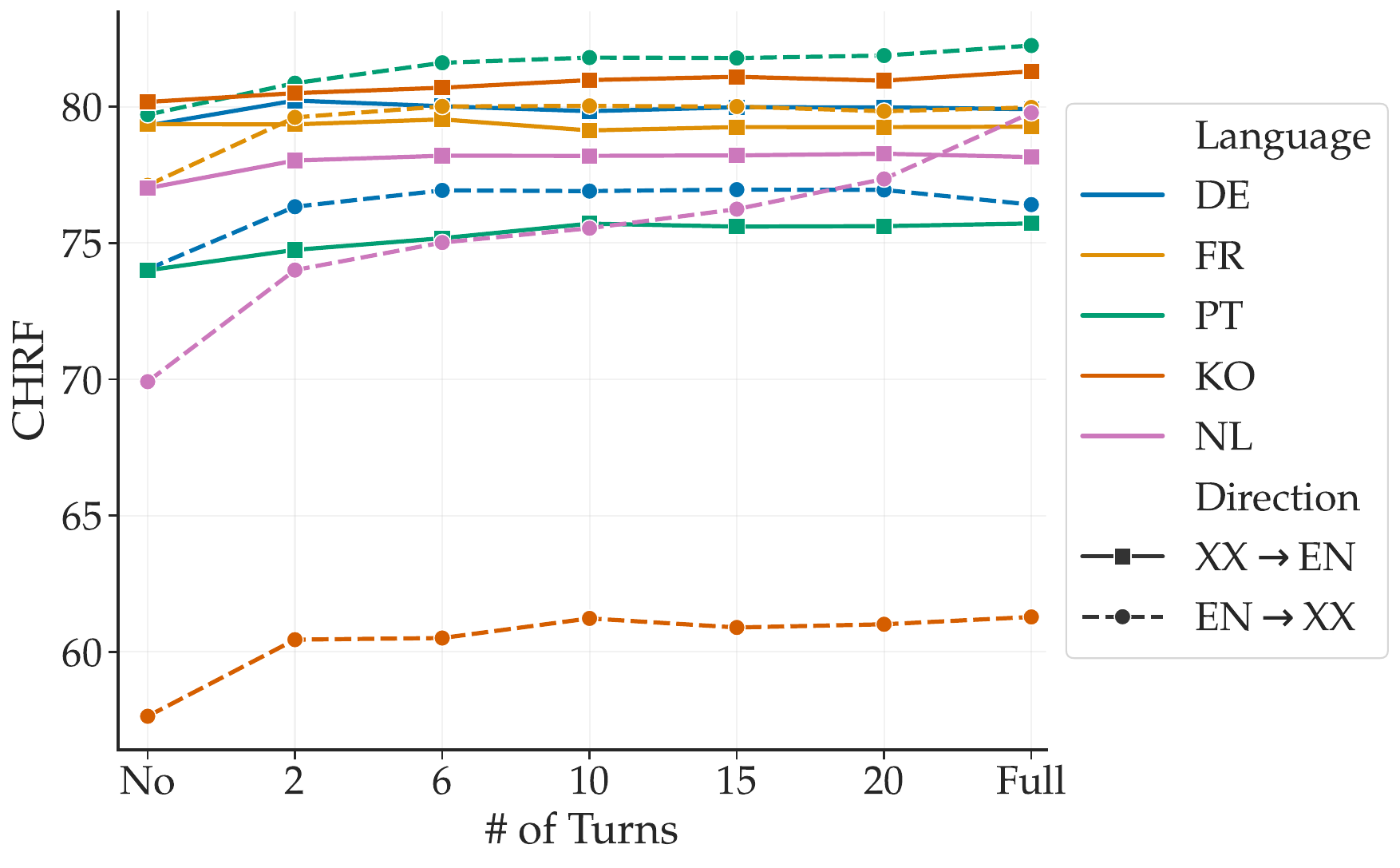} 
    \caption{Comparison of \chrf{} and \comet{} scores for varying context window sizes.}
    \label{fig:turn-analysis}
    % \vspace{-0.5cm}
\end{figure}

\begin{figure*}[t]
    \centering
    \begin{subfigure}[b]{0.45\linewidth}
        \centering
        \includegraphics[width=\linewidth,trim={0 2cm  0 0},clip]{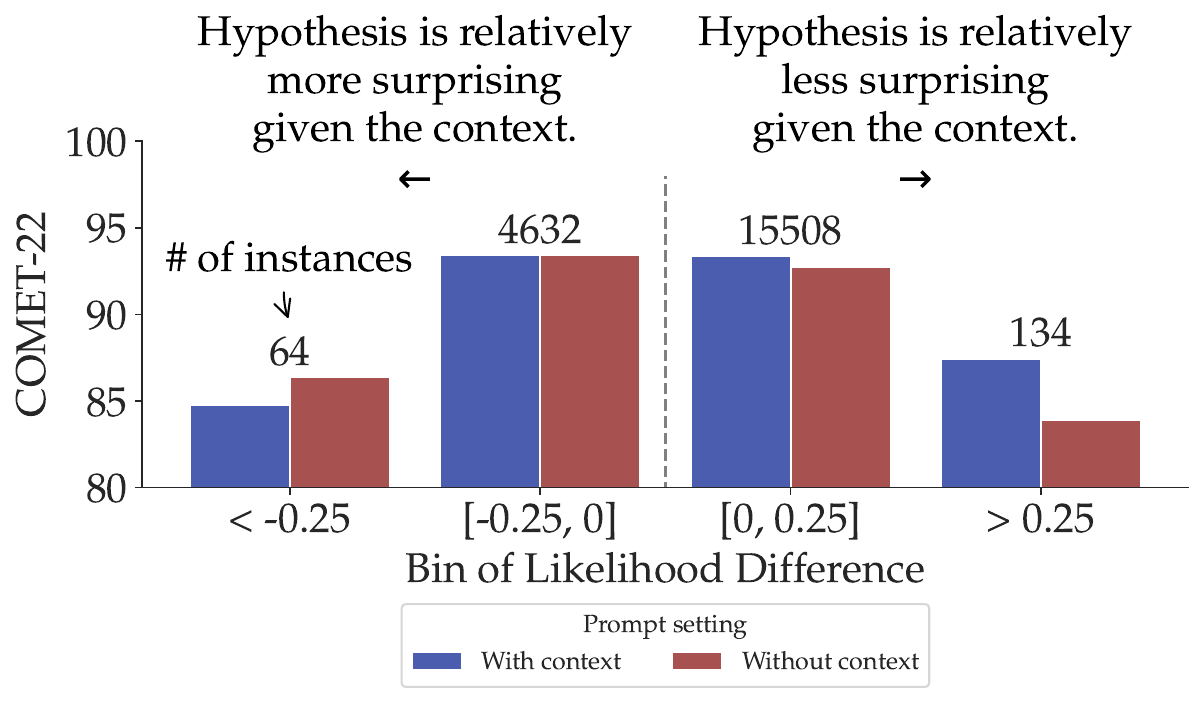}
        \caption{\TowerChat{}}
        \label{fig:pcxmi-hyp-a}
    \end{subfigure}
    % \vspace{1em}
    \begin{subfigure}[b]{0.45\linewidth}
        \centering
        \includegraphics[width=\linewidth,trim={0 2cm 0 0},clip]{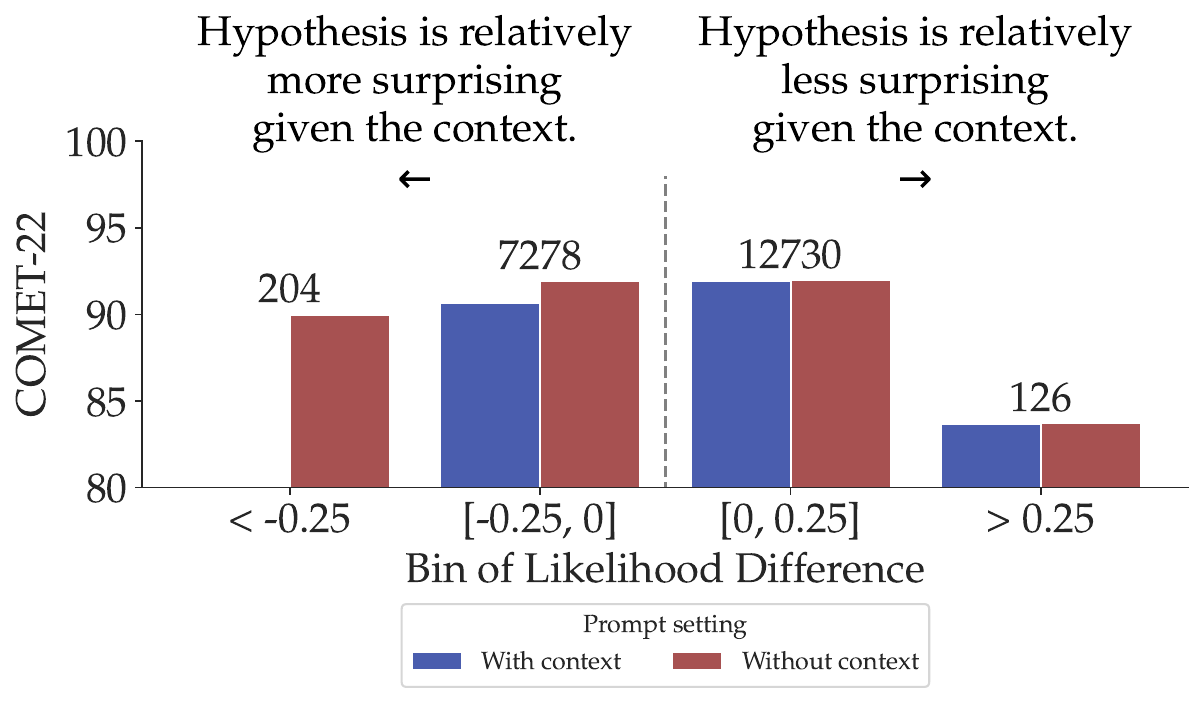}
        \caption{\TowerInstruct{}}
        \label{fig:pcxmi-hyp-b}
    \end{subfigure}
    % \vspace{-1\baselineskip}
    \caption{\comet{} under different prompt settings (\textcolor{CustomBlue}{with} and \textcolor{CustomRed}{without} context) for different bins of $\log \text{LD}$ on the hypothesis for \TowerChat{} (left) and \TowerInstruct{} (right).
    }
    \label{fig:pcxmi-hyp}
\end{figure*}

% \sweta{add that it implies one could use all context all the time}
Figure~\ref{fig:turn-analysis} shows the impact of varying the number of turns included in the context during inference on \comet{} and \chrf{}. For both metrics, translation quality improves as the context window length increases. Notably, for \textsc{en-nl} and \textsc{pt-en}, using the full context yields improvements beyond those achieved with up to 20 turns of conversation. In contrast, incorporating 6 to 10 turns is sufficient for other language pairs to reach peak performance. Furthermore, we observe limited gains when adding context for generating translations into English, consistent with prior observations \cite{agrawal2024context}. These results suggest that the optimal context window varies by language pair and that adaptive strategies for optimal context selection may be useful.

\subsection{How does context influence predictions?}

To understand when context meaningfully influences translations, we employ measures that quantify the impact of context on model predictions:
\begin{itemize}
    \item \textbf{P-CXMI}~\citep{fernandes-etal-2023-translation} measures how likely the \textit{reference} (contextual) translation $y$ is given a context, $C$, compared to when the context is not provided.
    \begin{equation}
     \text{P-CXMI} = \log \frac{P (y | x, C)}{P (y | x) } 
    \end{equation}
    Intuitively, a higher P-CXMI means that the reference translation requires context to be translated accurately. 
    \item \textbf{Likelihood Difference} \cite{shi-etal-2024-trusting} We measure the difference between the log-likelihood of a context-aware hypothesis against a context-agnostic hypothesis as:
\begin{equation}
    \log \text{LD} = \log \frac{P (\hat{y}_\text{ctx} | x, C)}{ P (\hat{y}_\text{no-ctx} | x) }
\end{equation}
   Unlike P-CXMI, this metric does not rely on a reference translation. Instead, it directly evaluates how incorporating context affects the likelihood of the generated hypothesis.

\end{itemize}
Our findings on how these metrics relate to translation quality are presented below:

\begin{figure}[t]
    \centering
    \includegraphics[width=\columnwidth, trim={0 1.9cm 0 0},clip]{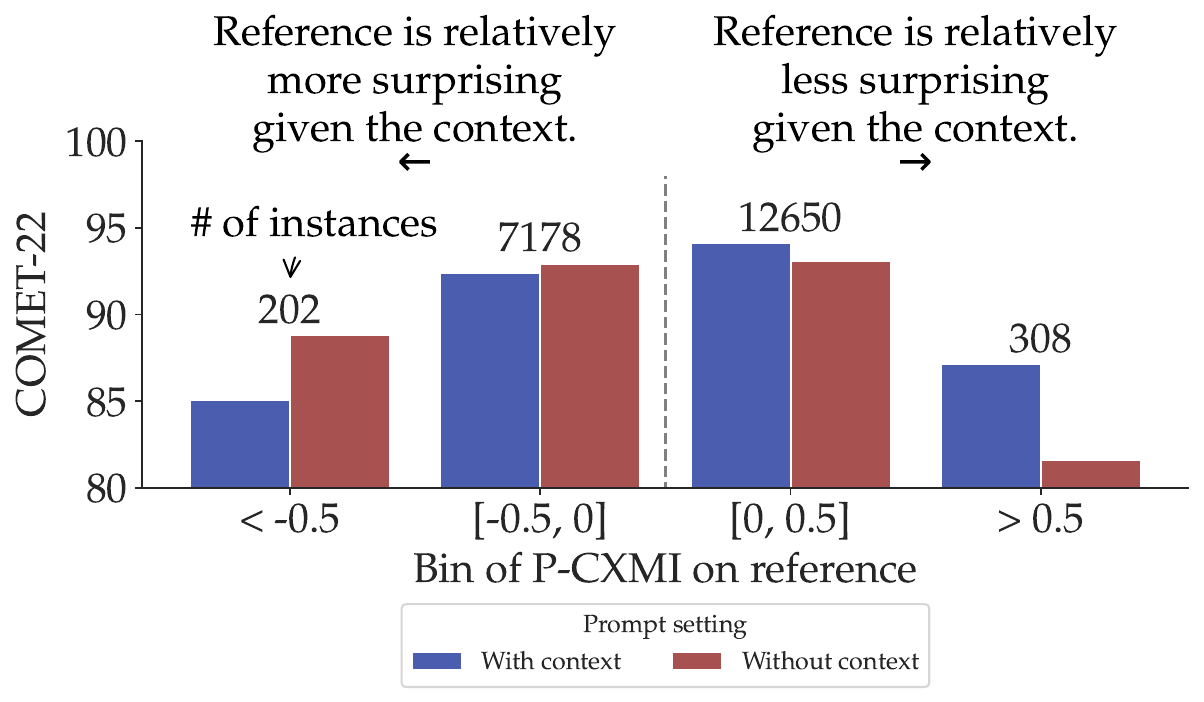}
    \caption{\comet{} under different prompts (\textcolor{CustomBlue}{with} and \textcolor{CustomRed}{without} context) for different P-CXMI bins. 
    }
    \label{fig:pcxmi-ref}
\end{figure}

\paragraph{Quality is higher for sentences that require resolving ambiguity by P-CXMI.}
Figure~\ref{fig:pcxmi-ref} shows that \TowerChat{} performs better with context when P-CXMI is positive and worse otherwise.
In other words, hypotheses are better on average when a reference translation requires context according to the model. 
The higher/lower the P-CXMI, the more positive/negative the change in quality.

\paragraph{The output likelihoods of \TowerChat{} predict the impact of context on quality.}
Figure~\ref{fig:pcxmi-hyp-a} shows that translation quality is higher for the context-aware hypothesis when it is more likely than the context-unaware hypothesis and vice-versa.
This means that we can predict, to a certain extent, whether a translation will benefit from context when using \TowerChat{}.
Remarkably, this does not hold for \TowerInstruct{}~(Figure~\ref{fig:pcxmi-hyp-b}): regardless of likelihoods, translation quality is always highest when not leveraging context.\footnote{Except for a small difference in the last bin of likelihood difference, this is also the case for the version of \TowerInstruct{} trained on context-unaware chat data (see Figure~\ref{fig:lhood-ablation}).}

\begin{figure}[t]
    \centering
    % \vspace{-1.5cm}
    \includegraphics[trim={37cm 36cm 37cm 36cm},clip,width=\linewidth]{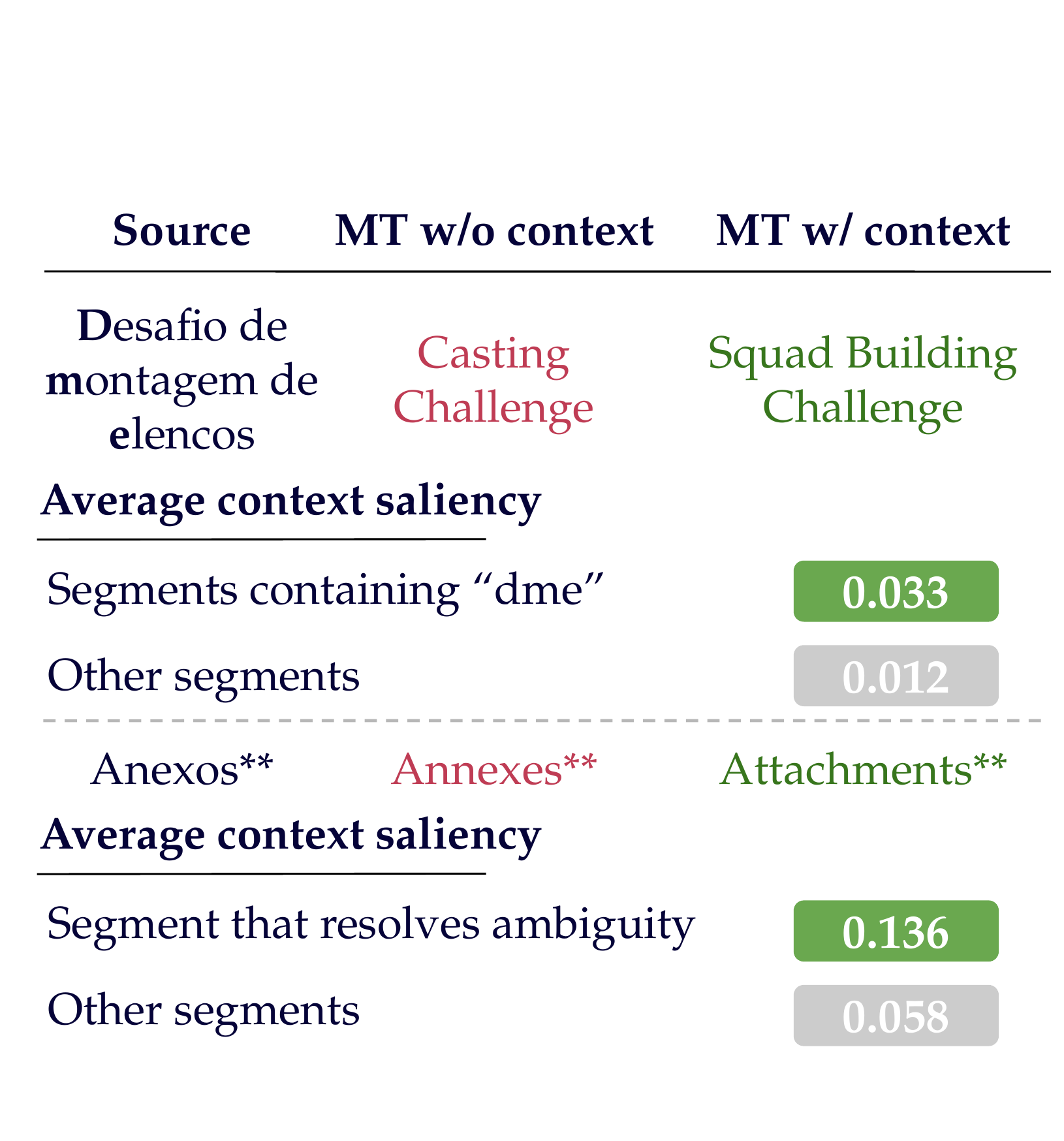}
    \caption{Two examples of PT$\rightarrow$EN  translations with and without context where the contextually-informed translation is accurate, while the translation without context is lexically correct but contextually incorrect: saliency values are high for context segments that resolve semantic ambiguity.
    }
    \label{fig:context-saliency}
    % \vspace{-0.5cm}
\end{figure}

\subsection{Which parts of the context impact target predictions?}

\paragraph{PeCoRE} \cite{sarti2024quantifying} is a method to identify salient context input tokens that explain the generation of context-sensitive output tokens over alternatives. Following their notation, we use $X_{ctx}: \{X, C\}$ and $X_{no-ctx}: X$ to represent the context-aware and context-agnostic inputs. These inputs are passed through \TowerChat{} to generate context-sensitive $\hat{y}_{ctx}$ and context-agnostic $\hat{y}_{no-ctx}$ outputs respectively. At each decoding step $i$, Let $P_{\text{ctx}}^i$ represent the token-level probability distribution obtained from the model when conditioned on $\{X_{ctx}, \hat{y}_{ctx, <i}\}$. To form a contrastive target distribution, we force-decode the non-contextual output, $y_{no-ctx}$ with the context-aware input, $X_{ctx}$, resulting in the probability distribution, $P_{\text{no-ctx}}^i$. The degree to which individual context tokens contribute to the contextualized generation is quantified by taking the gradient of a target function ($f_{tgt}$) with respect to each input token. We use the likelihood ratio between $P_{\text{ctx}}^i$ and $P_{\text{no-ctx}}^i$ as $f_{tgt}$.

\paragraph{Salient tokens signal important parts of context leveraged by \TowerChat{}.}

% \sweta{add a reference to appendix and also add en-ko examples}
In Figure~\ref{fig:context-saliency} we give two examples of translations produced by \TowerChat{} with and without context. The former is correct, while the latter is lexically correct but semantically incorrect.
We measure the saliency of segments in the context crucial to resolving the source text's semantical ambiguity.
\TowerChat{} shows high -- almost threefold -- saliency for these segments compared to the rest.
In the first case, ``dme'' is an acronym for the source text. It co-occurs with segments that make it clear that the conversation is about squads and not casts; \TowerChat{} with context is able to pick up on this and produce a correct translation.
In the second case, a segment in the context makes it obvious that ``Anexos'' should be translated to ``Attachments'' (as in email) rather than ``Annexes''. This segment is much more salient than the rest.
This shows that the model can correctly use the context it is provided with to generate correct translations in both lexical and contextual senses. We provide some additional examples along with the complete conversation context for the ones presented here in Appendix~\ref{app:examples}.

\section{Conclusion and Future Work}
This work presents a context-aware framework for improving translation quality in bilingual conversations. Experiments on two task-oriented domains show that the resulting model is better at leveraging contextual information during training and inference. Quality-aware decoding methods with hypotheses generated by a context-aware model further improve translation quality and accuracy in modeling discourse phenomena over strong baselines for both domains and all language pairs.

However, several challenges remain. 
% Our extensive analysis shows that better context-aware metrics are needed to capture the nuances of contextual relevance in translations.
% 
Our results indicate that context benefits low-quality segments most, and adding context beyond a specific number of turns yields limited gains. This suggests that context should be selectively incorporated where it has the most significant impact. One can design a dynamic selection mechanism by being strategic about the number of turns included and/or using more effective quality estimation metrics for segment-level chat translation \---\, particularly those aligned with reference-based metrics like COMET-22. Such a mechanism would apply context only when the estimated quality of a segment falls below a certain threshold \cite{farinhas2025translate}, ensuring more efficient and targeted use of contextual information. We could also use such a metric to perform QE reranking, which scales linearly with the number of candidates instead of MBR decoding. This would optimize computational efficiency while preserving translation quality where it matters most. We leave a detailed investigation of these strategies to future work.
% Furthermore, while our analysis provides valuable insights into how the model leverages context, we need better evaluation frameworks to understand how context can be adapted to improve the model's attention to salient information. 

% \section*{Acknowledgments}

% We thank John Mendonça, Ben Peters, Giuseppe Attanasio, Miguel Ramos, Duarte Alves and the members of the SARDINE lab for their constructive feedback on the paper.
% This work was supported by EU's Horizon Europe Research and Innovation Actions (UTTER, contract 101070631), by the project DECOLLAGE (ERC-2022-CoG 101088763), by the Portuguese Recovery and Resilience Plan through project C645008882-00000055 (Center for Responsible AI), and by Fundação para a Ciência e Tecnologia through contract UIDB/50008/2020.

\bibliography{anthology_0,anthology_1,custom}
\bibliographystyle{acl_natbib}

% \iftaclpubformat

\onecolumn

\appendix
\section{Context Prompt Example}\label{appendix:context-prompt-example}
\vspace{-0.3cm}
\begin{figure}[h]
         \small
         \renewcommand\tabularxcolumn[1]{m{#1}}% <-- added
        \renewcommand\arraystretch{1.2}
        \centering
    \begin{tabularx}{0.95\linewidth}{*{1}{>{\arraybackslash}X}}
Context: Naja es geht so. \\
 Ich habe gestern einen ärgerlichen Vorfall. \\
 Ich hatte auf meinem ACC knapp 335000 PRS-ORG Coins und beim anmelden hatte Ich nur noch 776\\
 So you're missing your coins. \\
 That's indeed concerning. \\
 And I'll surely look into this. \\
 Please provide me the email of the account. \\
Thank you. \\
Let me check if there were any transaction for coins that were not done by you. \\
Thank you. \\
I can see there are no suspicious activity on your account in past 7 days. \\
 I can see all the coins were used by your web app and PRS-ORG. \\ \\
   Translate the English source text to German, \textcolor{purple}{given the context.} \\
   English: Let me tell you where it was used. \\
   German: \textcolor{CustomBlue}{Lassen Sie mich Ihnen sagen, wo es verwendet wurde.} \\
   \bottomrule
    \end{tabularx}  
\caption{Specific training instance with context for \TowerChat{}. Gradient updates are only performed on the \textcolor{CustomBlue}{reference}, and new lines are encoded as \textbackslash{}n.} \label{fig:prompt_context_concrete}
\vspace{-0.4cm}
\end{figure}

\section{\muda-based Evaluation} \label{app:muda}
% We use \muda to evaluate whether the model correctly handles specific discourse phenomena. 
\muda{} uses a series of rule-based taggers to identify tokens in reference and hypothesis translations that are related to specific discourse phenomena. These tokens often correspond to tokens where discourse coherence is particularly challenging, such as pronouns (e.g., ``it'' in English, which could translate to ``er,'' ``sie,'' or ``es'' in German depending on the antecedent’s gender). Native speakers verify the tagging rules to ensure linguistic validity. We then compute the F1 score for each tag based on whether a tagged word in the reference/hypothesis also appears and is tagged in the hypothesis/reference.
% For example:

\section{\TowerChat{} Training Hyperparameters}\label{appendix:towerchat-hyperparameters}
We do full finetuning of \TowerBase{} with the hyperparameters of \TowerInstruct{}~\citep{alves2024tower} (see Table~\ref{tab:towerchat-hyperparameters}).

\begin{table}[ht]
\renewcommand{\arraystretch}{1.1}
\begin{center}
\begin{tabular}{ll}
\toprule
Precision & bfloat16 \\
Packing & True \\
Global train batch size & 256 \\
Number of Epochs & 4 \\
Learning rate & 7e-6 \\
LR Scheduler & cosine \\
Warmup Steps & 500 \\
Weight Decay & 0.01 \\
Optimizer & Adam \citep{Kingma2014AdamAM} \\
Adam ($\beta_1$, $\beta_2$, $\epsilon$)  & (0.9, 0.999, 1e-8) \\
Maximum Sequence Length & 2048 \\
\bottomrule
\end{tabular}    
\end{center}
\caption{Hyperparameter configuration.}
\label{tab:towerchat-hyperparameters}
\end{table}

\section{Oracle Analysis - \chrf{}} 
\label{sec:oracle}
We present an oracle analysis on the quality of hypotheses generated by the \TowerChat{} and \TowerInstruct{} as measured by \chrf{} in Table~\ref{tab:oracle_chrf}: the candidates generated by \TowerChat{} have a higher overlap with the reference than those generated by \TowerInstruct{}.

% \vspace{-0.3cm}
\begin{table}[h]
    \centering
    \resizebox{0.90\linewidth}{!}{
    \begin{tabular}{lrrrcrrr}
    \toprule
        LP & \multicolumn{3}{c}{EN-XX} & \multicolumn{3}{c}{XX-EN}\\
        & \TowerChat{} & \TowerInstruct{} & $\Delta$ && \TowerChat{} & \TowerInstruct{} & $\Delta$ \\

         \midrule 
         DE & 90.21 & 88.92 & +1.29 & & 91.77 & 92.92 & -1.15 \\ 
         FR & 90.65 & 89.89 & +0.76 & & 93.19 & 94.91 & -1.72 \\ 
         KO & 83.18 & 67.15 & +16.03 & & 93.56 & 92.15 & +1.41 \\ 
         NL & 91.82 & 83.69 & +8.13 & & 91.93 & 91.25 & +0.68 \\ 
         PT & 94.04 & 87.03 & +7.01 & & 91.99 & 86.30 & +5.69 \\ 
         \bottomrule
    \end{tabular}
    }
    \caption{Oracle \chrf{} scores on the pool of candidates generated by the two configurations: \TowerChat{} with context-aware prompt and \TowerInstruct{} with context-agnostic prompts, respectively.}
    \label{tab:oracle_chrf}
    % \vspace{-0.4cm}
\end{table}

\section{Test Results and Analysis by Language Pair} \label{appendix-analysis-lps}

% \sweta{add missing results}

Tables \ref{tab:main_results_lps_chrf}, \ref{tab:main_results_lps_comet} and \ref{tab:main_results_lps_metricx} show \chrf{}, \comet{} and \metricx{} scores for individual language pairs across evaluated settings.

\begin{table*}[htb!]
    \centering
    \footnotesize
    \resizebox{0.90\linewidth}{!}{
    \begin{tabular}{lcrrrrrcrrrrr}
    \toprule
    & &  \multicolumn{5}{c}{\textbf{\textsc{en-xx}}} && \multicolumn{5}{c}{\textbf{\textsc{xx-en}}} \\
    \cline{3-7} \cline{9-13}
    \addlinespace[0.1cm]
    \multirow{-2}{*}{\textbf{\textsc{Model}}} &  \multirow{-2}{*}{\textbf{\textsc{Context?}}} & \multicolumn{1}{c}{\textsc{de}} & \multicolumn{1}{c}{\textsc{fr}}  & \multicolumn{1}{c}{\textsc{pt}} & \multicolumn{1}{c}{\textsc{ko}} & \multicolumn{1}{c}{\textsc{nl}} && \multicolumn{1}{c}{\textsc{de}} & \multicolumn{1}{c}{\textsc{fr}}  & \multicolumn{1}{c}{\textsc{pt}} & \multicolumn{1}{c}{\textsc{ko}} & \multicolumn{1}{c}{\textsc{nl}} \\ 
    \midrule   
   \multicolumn{13}{l}{\small \bf Baselines} \\

\gptfouro  & \xmark  &76.67 & 78.69 & 75.81 & 47.63 & 71.65 && 79.60 & 78.27 & 73.73 & 76.45 & 78.62   \\
& \cmark & 74.96 & 78.45 & 76.05 & 48.78 & 73.49 && 78.01 & 77.64 & 72.58 & 69.21 & 76.28   \\
  \cdashlinelr{1-13}
  
 \textbf{ \TowerInstruct{} } & \xmark  &  71.81 & 74.59 & 72.26 & 43.18 & 62.90 && 77.57 & 79.02 & 72.06 & 75.73 & 75.80\\
& \cmark  &71.16 & 74.38 & 68.50 & 41.70 & 61.23 && 75.68 & 78.31 & 71.83 & 72.63 & 73.15 \\
     \qquad + QAD (\comet{}) & \xmark  & 72.74 & 73.60 & 72.95 & 43.11 & 63.59 && 76.26 & 79.44 & 71.86 & 75.32 & 75.07  \\
    \qquad + QAD (\contextcomet{}) & \xmark &  72.05 & 74.08 & 72.96 & 43.19 & 63.02 && 76.80 & 79.13 & 72.15 & 75.88 & 75.61 \\ 
       
        \cdashlinelr{1-13}
  
   \textbf{\TowerChat{}}& \xmark & 74.04 & 77.12 & 79.71 & 57.63 & 69.91 && 79.31 & 79.36 & 74.00 & 80.17 & 77.01\\
   & \cmark &76.41 & 79.97 & 82.24 & 61.28 & 79.78 && 79.91 & 79.26 & 75.72 & 81.30 & 78.15  \\ 
   & \hspace{0.45cm} \cmark (en) & 77.31 & 80.27 & 81.67 & 61.06 & 73.73 && 79.87 & 79.14 & 74.97 & 81.52 & 78.31  \\
    \qquad + QAD (\comet{})  & \cmark &77.09 & 80.34 & 82.25 & 61.79 & \textbf{80.33} && 79.70 & \textbf{78.78} & 75.88 & 81.56 & \textbf{78.67}  \\
    \qquad + QAD (\contextcomet{})  &\cmark & \textbf{77.23} & \textbf{80.51} & \textbf{82.55} & \textbf{62.29} & 80.25 && \textbf{79.87} & 78.57 & \textbf{76.01} & \textbf{81.57} & 78.60   \\

   \cdashlinelr{1-13}
    
    \qquad + SFT on QAD (\contextcomet{}) & \xmark& 72.30 & 77.25 & 79.74 & 58.46 & 69.74 && 78.52 & 79.78 & 73.66 & 80.49 & 77.17 \\
    & \cmark & 76.36 & 79.67 & 81.82 & 61.09 & 78.62 && 79.01 & 78.52 & 75.64 & 80.52 & 77.33\\
    \bottomrule
    \end{tabular}}
    \caption{Results by \chrf{} ($\uparrow$) on WMT24 Chat Shared Task Dataset by Language Pair. }
    \label{tab:main_results_lps_chrf}
\end{table*}

\begin{table*}[htb!]
    \centering
    \footnotesize
    \resizebox{0.90\linewidth}{!}{
    \begin{tabular}{lcrrrrrcrrrrr}
    \toprule
    & &  \multicolumn{5}{c}{\textbf{\textsc{en-xx}}} && \multicolumn{5}{c}{\textbf{\textsc{xx-en}}} \\
    \cline{3-7} \cline{9-13}
    \addlinespace[0.1cm]
    \multirow{-2}{*}{\textbf{\textsc{Model}}} &  \multirow{-2}{*}{\textbf{\textsc{Context?}}} & \multicolumn{1}{c}{\textsc{de}} & \multicolumn{1}{c}{\textsc{fr}}  & \multicolumn{1}{c}{\textsc{pt}} & \multicolumn{1}{c}{\textsc{ko}} & \multicolumn{1}{c}{\textsc{nl}} && \multicolumn{1}{c}{\textsc{de}} & \multicolumn{1}{c}{\textsc{fr}}  & \multicolumn{1}{c}{\textsc{pt}} & \multicolumn{1}{c}{\textsc{ko}} & \multicolumn{1}{c}{\textsc{nl}} \\ 
    \midrule   
   \multicolumn{13}{l}{\small \bf Baselines} \\
    % \nllb{} & \xmark & 90.56 & 91.06 & 86.33 & 87.26 & 87.86 && 89.03 & 89.18 & 86.1 & 88.05 & 88.45 \\
\gptfouro  & \xmark  & 92.74 & 92.43 & 93.01 & 92.26 & 92.68 && 92.16 & 92.18 & 91.40 & 93.24 & 93.07 \\

& \cmark & 92.49 & 92.62 & 93.40 & 93.06 & 93.08 && 91.95 & 91.76 & 91.19 & 90.67 & 92.39   \\
  \cdashlinelr{1-13}
  
 \textbf{ \TowerInstruct{} } & \xmark  & 91.71 & 91.89 & 91.90 & 91.64 & 91.30 && 92.08 & 92.78 & 90.43 & 93.13 & 92.45 \\
& \cmark  & 91.48 & 91.08 & 90.79 & 91.13 & 91.00 && 91.33 & 91.89 & 90.63 & 91.88 & 91.08 \\
& \hspace{0.45cm} \cmark (en) & 92.94 & 92.70 & 94.38 & 94.10 & 92.51 && 92.30 & 92.68 & 91.67 & 94.87 & 92.98 \\
     \qquad + QAD (\comet{}) & \xmark  & 92.77 & 92.62 & 93.24 & 93.04 & 92.68 && \textbf{92.78} & \textbf{93.34} & 91.13 & 93.87 & 92.88  \\
    \qquad + QAD (\contextcomet{}) & \xmark & 92.53 & 92.49 & 92.86 & 92.75 & 92.23 && 92.58 & 93.2 & 90.87 & 93.81 & 92.77 \\ 
        \cdashlinelr{1-13}
 % \TowerInstruct{} 7B  (Chat) & \xmark & 92.86 & 92.59 & 94.37 & 93.92 & 93.94 && 92.18 & 92.93 & 91.44 & 94.57 & 92.89   \\
 %  & \cmark & 92.01 & 91.99 & 93.50 & 92.96 & 93.16 && 90.92 & 91.33 & 90.88 & 92.01 & 90.90\\
   \textbf{  \TowerChat{}}& \xmark & 92.36 & 92.26 & 93.89 & 93.73 & 92.81 && 92.28 & 92.79 & 91.06 & 94.69 & 92.78 \\
   & \cmark & 92.74 & 92.64 & 94.53 & 94.13 & 94.09 && 92.24 & 92.67 & 92.09 & 94.98 & 93.06 \\ 
    \qquad + QAD (\comet{})  & \cmark & \textbf{93.28} & \textbf{93.13} & \textbf{94.91} & \textbf{95.01} & \textbf{94.54} && 92.58 & 92.95 & \textbf{92.63} & \textbf{95.32} & \textbf{93.49}  \\
    \qquad + QAD (\contextcomet{})  & \cmark & 93.22 & 92.96 & 94.76 & 94.96 & 94.36 && 92.48 & 92.71 & 92.46 & 95.16 & 93.38  \\
      \cdashlinelr{1-13}
    
    \qquad + SFT on QAD (\contextcomet{}) & \xmark & 92.55 & 92.68 & 93.96 & 93.96 & 93.10 && 92.34 & 92.93 & 91.18 & 94.71 & 92.87\\
    & \cmark& 92.96 & 92.84 & 94.39 & 94.40 & 94.02 && 92.27 & 92.69 & 92.13 & 95.05 & 93.04  \\
    \bottomrule
    \end{tabular}}
    \caption{Results by \comet{} ($\uparrow$) on WMT24 Chat Shared Task Dataset by Language Pair. }
    \label{tab:main_results_lps_comet}
\end{table*}

\begin{table*}[htb!]
    \centering
    \footnotesize
    \resizebox{0.90\linewidth}{!}{
    \begin{tabular}{lcrrrrrcrrrrr}
    \toprule
    & &  \multicolumn{5}{c}{\textbf{\textsc{en-xx}}} && \multicolumn{5}{c}{\textbf{\textsc{xx-en}}} \\
    \cline{3-7} \cline{9-13}
    \addlinespace[0.1cm]
    \multirow{-2}{*}{\textbf{\textsc{Model}}} &  \multirow{-2}{*}{\textbf{\textsc{Context?}}} & \multicolumn{1}{c}{\textsc{de}} & \multicolumn{1}{c}{\textsc{fr}}  & \multicolumn{1}{c}{\textsc{pt}} & \multicolumn{1}{c}{\textsc{ko}} & \multicolumn{1}{c}{\textsc{nl}} && \multicolumn{1}{c}{\textsc{de}} & \multicolumn{1}{c}{\textsc{fr}}  & \multicolumn{1}{c}{\textsc{pt}} & \multicolumn{1}{c}{\textsc{ko}} & \multicolumn{1}{c}{\textsc{nl}} \\ 
    \midrule   
   \multicolumn{13}{l}{\small \bf Baselines} \\

\gptfouro  & \xmark  & 0.42 & 0.28 & 0.29 & 0.51 & 0.33 && 0.52 & 0.51 & 0.67 & 0.32 & 0.46  \\
& \cmark & 0.45 & 0.22 & 0.28 & 0.39 & 0.30 && 0.55 & 0.56 & 0.64 & 0.65 & 0.50   \\
  \cdashlinelr{1-13}
  
 \textbf{ \TowerInstruct{} } & \xmark & 0.28 & 0.23 & 0.43 & 0.57 & 0.37 && 0.50 & 0.53 & 0.86 & 0.37 & 0.53 \\
& \cmark  & 0.38 & 0.29 & 0.69 & 0.60 & 0.49 && 0.56 & 0.55 & 0.74 & 0.46 & 0.69 \\
     \qquad + QAD (\comet{}) & \xmark  & \textbf{0.25} & 0.20 & 0.37 & 0.42 & 0.31 && 0.48 & 0.49 & 0.78 & 0.35 & 0.50  \\
    \qquad + QAD (\contextcomet{}) & \xmark & 0.26 & \textbf{0.19} & 0.33 & 0.43 & 0.31 && 0.47 & 0.48 & 0.79 & 0.33 & 0.50  \\ 
       
        \cdashlinelr{1-13}
  
   \textbf{  \TowerChat{}}& \xmark & 0.27 & 0.24 & 0.29 & 0.42 & 0.37 && 0.50 & 0.51 & 0.71 & 0.33 & 0.52 \\
   & \cmark & 0.34 & 0.26 & 0.27 & 0.45 & 0.27 && 0.47 & 0.48 & 0.60 & 0.30 & 0.48 \\ 
   & \hspace{0.45cm} \cmark (en)  & 0.26 & 0.24 & 0.28 & 0.46 & 0.41 && 0.49 & 0.47 & 0.60 & 0.30 & 0.49\\
    \qquad + QAD (\comet{})  & \cmark & 0.30 & 0.22 & \textbf{0.24} & 0.31 & \textbf{0.21} && \textbf{0.46} & \textbf{0.46} & \textbf{0.55} & \textbf{0.27} & \textbf{0.45} \\
    \qquad + QAD (\contextcomet{})  &\cmark &  0.31 & 0.22 & \textbf{0.24} & \textbf{0.29} & 0.23 && 0.47 & 0.47 & 0.56 & \textbf{0.27} & \textbf{0.45}   \\
      \cdashlinelr{1-13}
    
    \qquad + SFT on QAD (\contextcomet{}) & \xmark& 0.29 & 0.23 & 0.26 & 0.38 & 0.37 && 0.50 & 0.50 & 0.73 & 0.33 & 0.52  \\
    & \cmark& 0.31 & 0.23 & 0.29 & 0.37 & 0.26 && 0.50 & 0.49 & 0.59 & 0.32 & 0.51  \\
    \bottomrule
    \end{tabular}}
    \caption{Results by \metricx{} ($\downarrow$) on WMT24 Chat Shared Task Dataset by Language Pair. }
    \label{tab:main_results_lps_metricx}
\end{table*}

\begin{figure}[htb!]
    \centering
    \includegraphics[width=0.90\columnwidth]{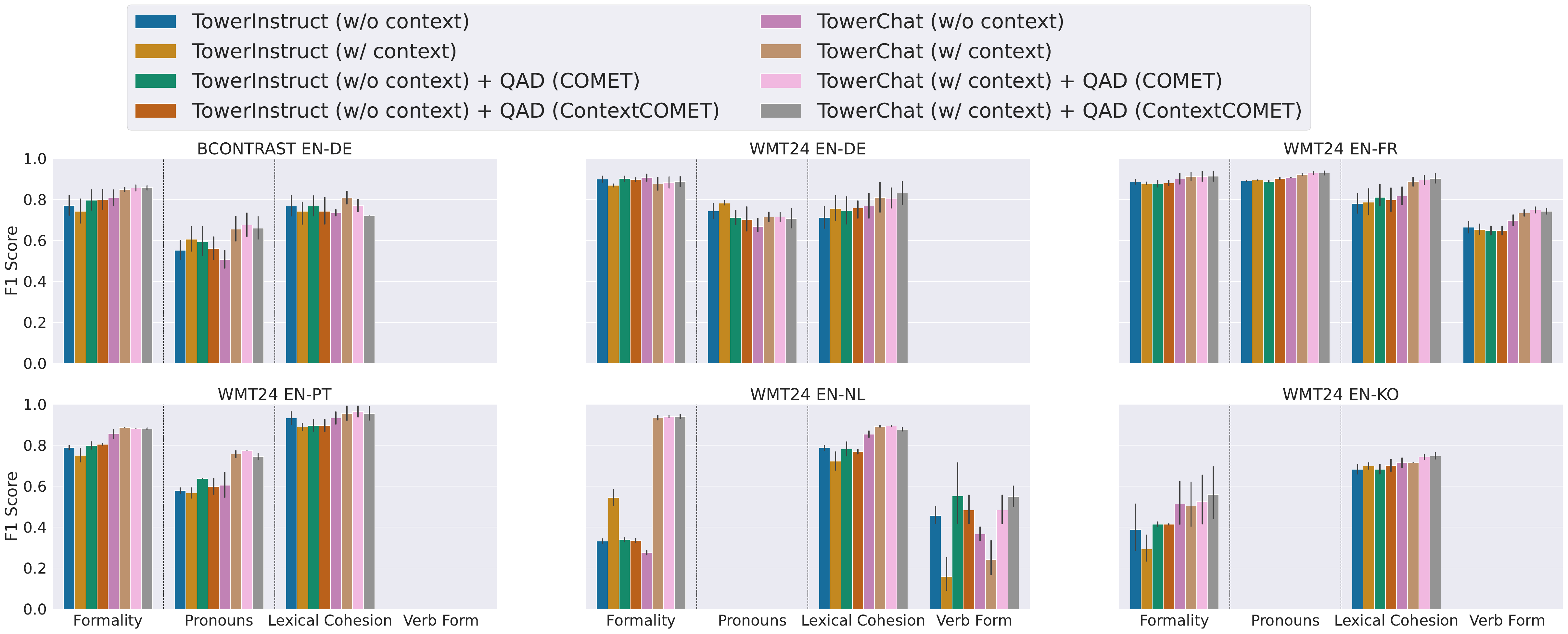}
    \caption{MuDA F1 by language pairs. LC: Lexical Cohesion, VF: Verb Form, P: Pronouns, F: Formality. On average, QAD with \contextcomet{} has the best F1 score in 7 out of 12 settings.}
    \label{fig:muda-accuracy-lps}
\end{figure}

\begin{figure}[htb!]
    \centering
    \includegraphics[width=0.45\columnwidth, trim={0 1.9cm 0 0},clip]{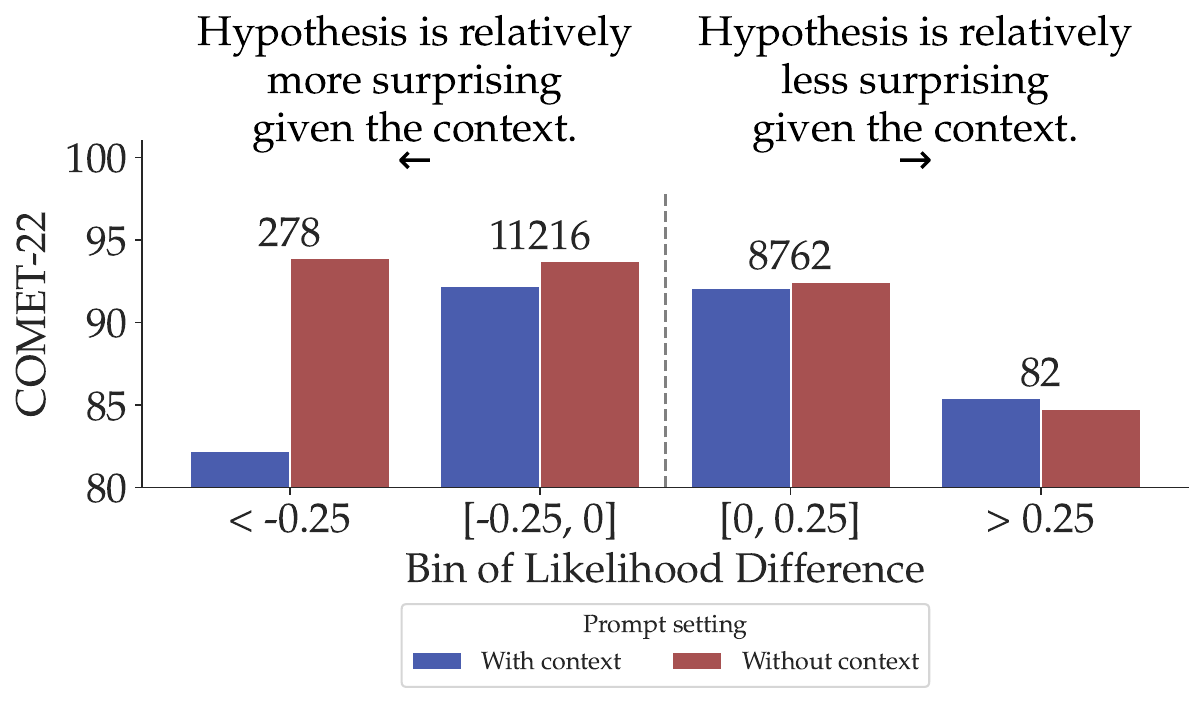}
    \caption{\comet{} under different prompts (\textcolor{CustomBlue}{with} and \textcolor{CustomRed}{without} context) for different P-CXMI bins for \TowerInstruct{} trained on chat domain data but context-unaware. 
    }
    \label{fig:lhood-ablation}
\end{figure}

\begin{figure*}[htb!]
    \centering
    \includegraphics[width=0.80\linewidth,trim={0 1.8cm 0 0},clip]{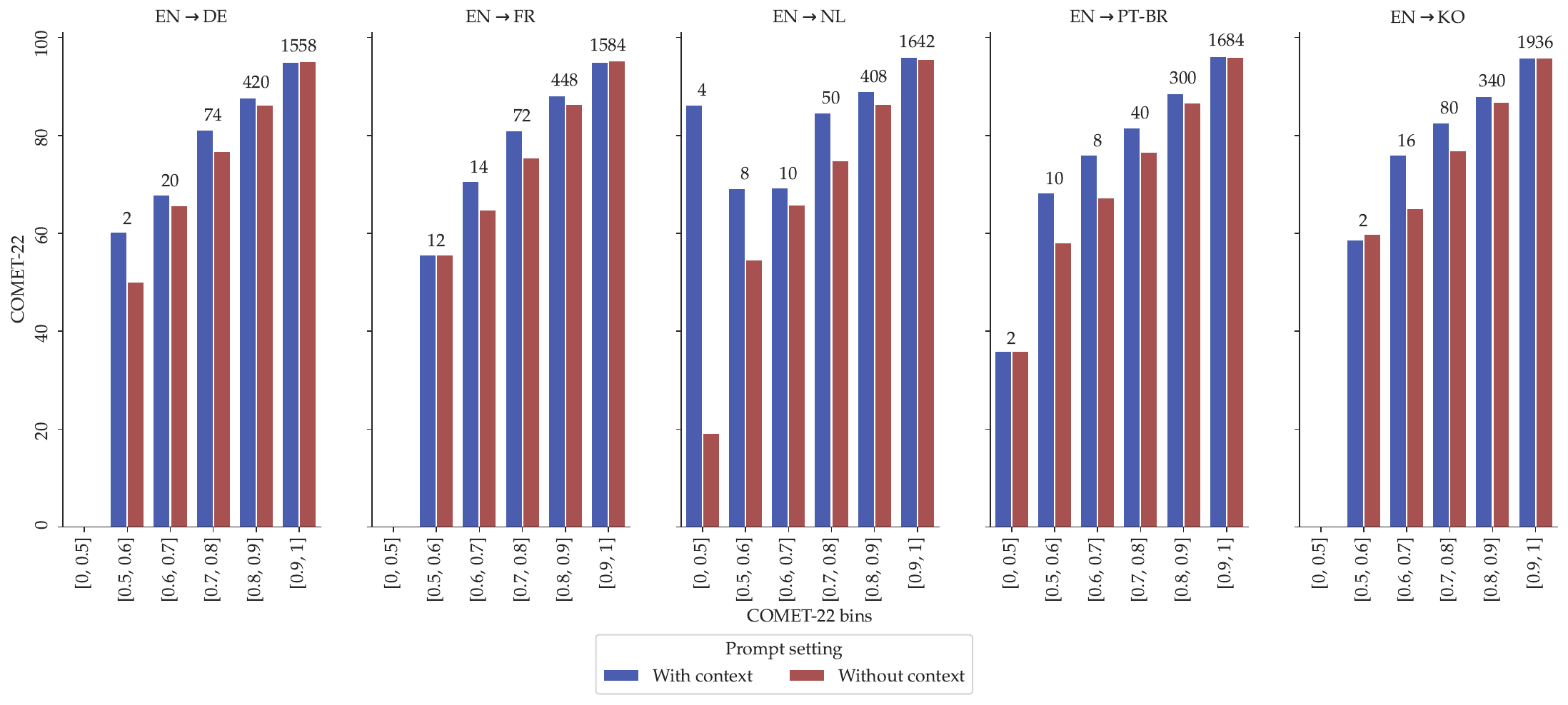}
     \includegraphics[width=0.80\linewidth,trim={0 1.8cm 0 0},clip]{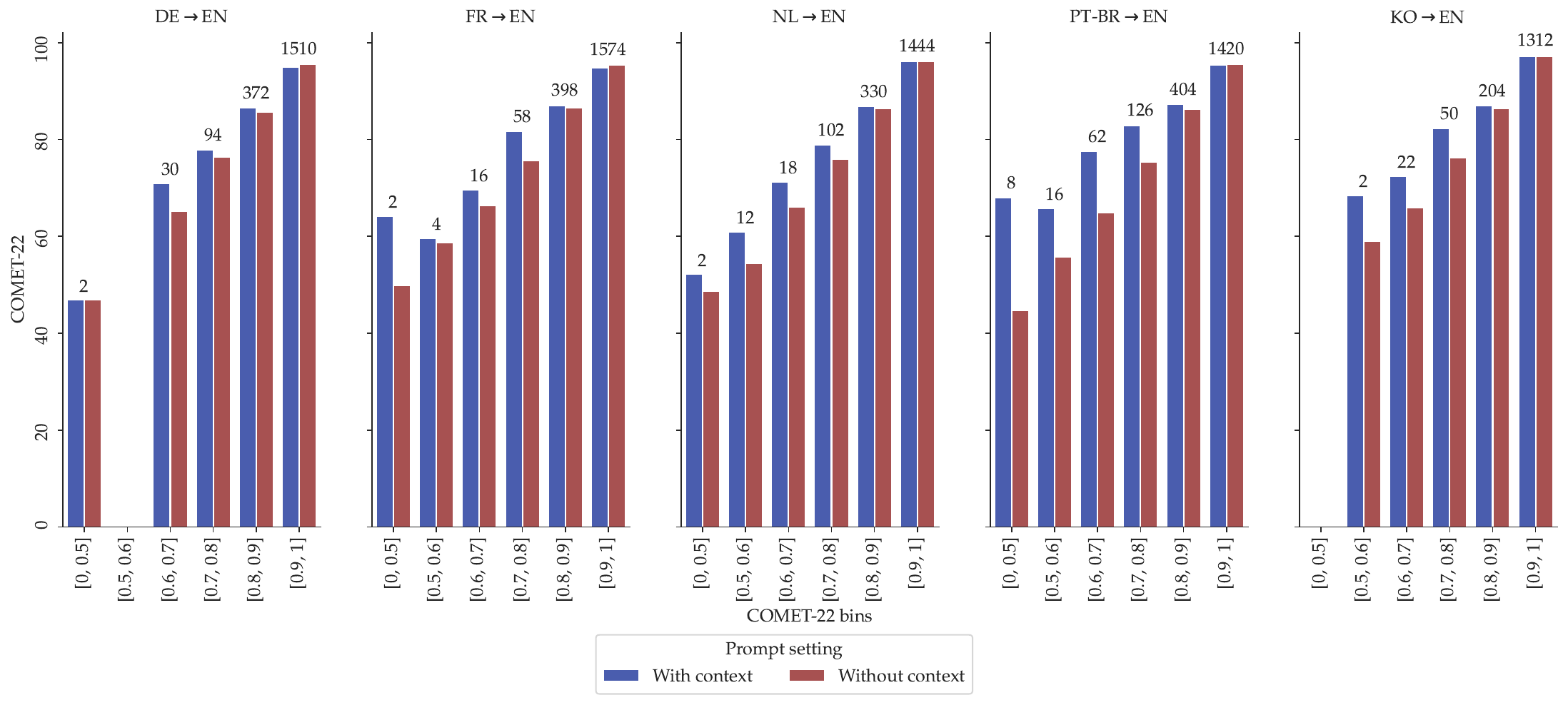}
    \caption{Quality Bins from \comet{} for \TowerChat{} \textit{w/o context} on EN$\rightarrow{}$XX (top) and XX$\rightarrow{}$EN (bottom) language pairs. \textcolor{CustomBlue}{Blue}: w/ context. \textcolor{CustomRed}{Red}: w/o context.
    }
    \label{fig:comet-bins-context-lps}
\end{figure*}

\begin{figure*}[htb!]
    \centering
    \includegraphics[width=0.80\linewidth,trim={0 1.8cm 0 0},clip]{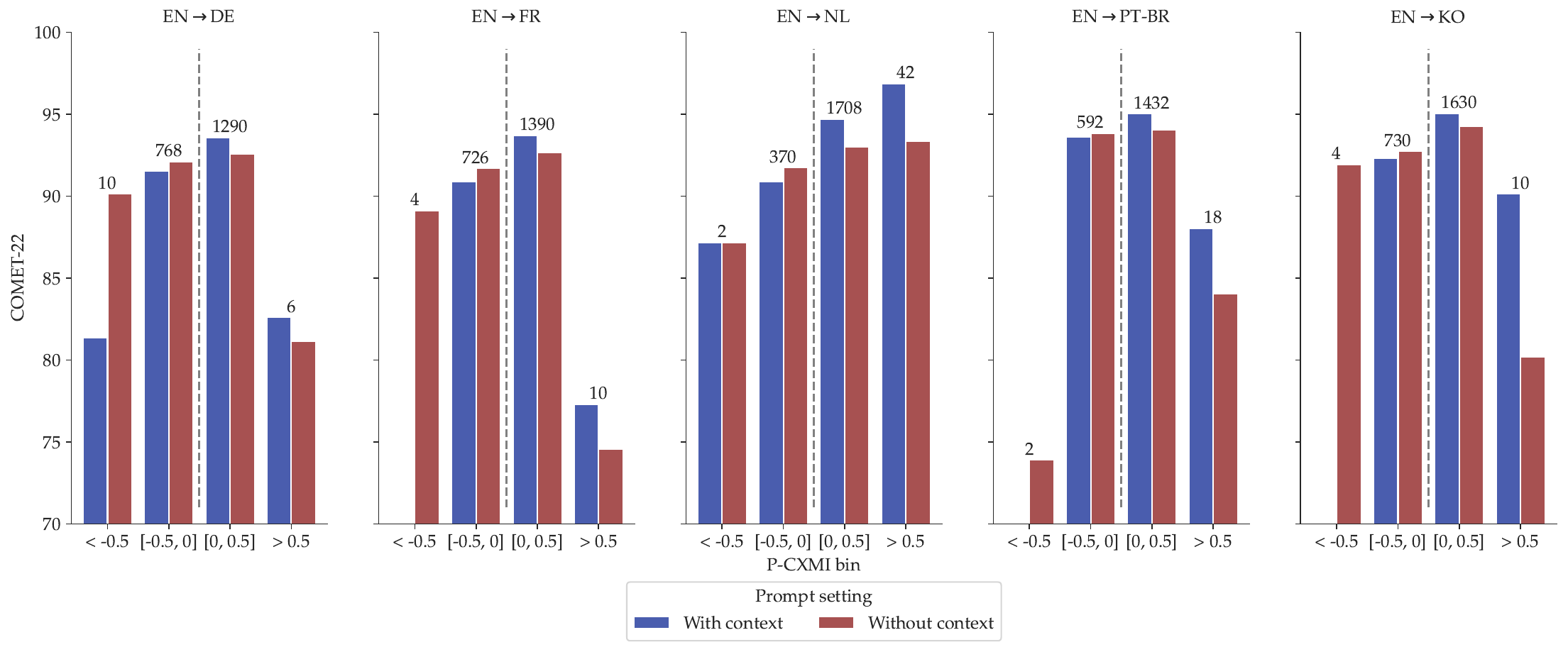}
     \includegraphics[width=0.80\linewidth,trim={0 1.7cm 0 0},clip]{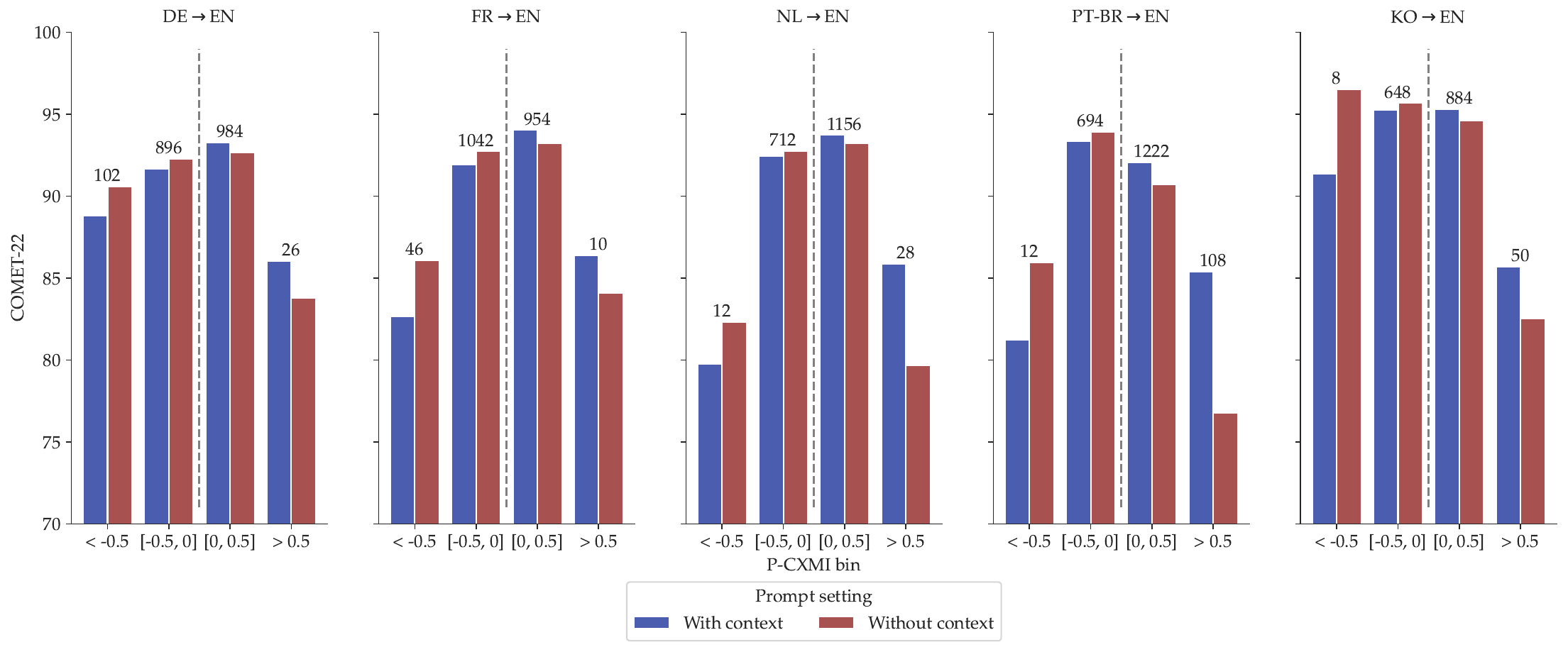}
    \caption{\comet{} under different prompt settings (\textcolor{CustomBlue}{with} and \textcolor{CustomRed}{without} context) for different P-CXMI bins and language pairs.
    }
    \label{fig:pcxmi-ref-lps}
\end{figure*}

% \newpage

\section{Additional Examples} \label{app:examples}
We show concrete examples where context results in a more contextually appropriate translation in Figures~\ref{fig:example_1},\ref{fig:example_2},\ref{fig:example_3} and \ref{fig:example_4}.

\begin{figure}[h]
         \small
         \renewcommand\tabularxcolumn[1]{m{#1}}% <-- added
        \renewcommand\arraystretch{1.2}
        \centering
    \begin{tabularx}{0.95\linewidth}{*{1}{>{\arraybackslash}X}}
Context: Obrigado por entrar em contato com a Ajuda PRS-ORG, o meu nome é NAME-N. \\
Para começar, pode-me indicar o seu nome por favor? \\
NAME-M \\
Hello NAME-M! \\
Its pleasure chat with you, how are you doing today? \\
Olá, tudo bem? \\ 
Gostaria de ajuda sobre o dme do PRS-ORG \\
Eu comecei a fazer o dme mas ele pede cartas que estão fora de packs e extintas no mercado \\
You are an incredible human! \\
Thank you so much for asking. \\ 
I am doing well and everything would be better if more people were like you. \\
Eu ja fiz 13 partes do dme e agora não consigo fazer o resto \\
Não acho justo com os jogadores manter um dme tão caro que seja impossível de fazer \\
Gostaria de receber os jogadores de volta ou receber os que faltam, no caso capitaes do PRS-ORG \\
Please do not worry, and be rest assured, I will try my best to resolve your issue regarding PRS-ORG \\
and cheer you up! \\
May I know what you referring with PRS-ORG? \\
I am sorry the term is not clear to me. \\ \\
   Translate the Brazilian Portuguese source text to English, given the context. \\
   Brazilian Portuguese: Desafio de montagem de elencos \\
   English: \\ \\
   \TowerChat{} w/o Context: \textcolor{BadRed}{Casting Challenge} \\
   \TowerChat{} w/ Context: \textcolor{GoodGreen}{Squad Building Challenge} \\
   \bottomrule
    \end{tabularx}  
\caption{Full conversation of first example in Figure~\ref{fig:turn-analysis}. In \textcolor{BadRed}{red}, the translation without context, which is wrong; in \textcolor{GoodGreen}{green}, the translation with context, which is correct.} \label{fig:example_1}
\end{figure}

\begin{figure}[h]
         \small
         \renewcommand\tabularxcolumn[1]{m{#1}}% <-- added
        \renewcommand\arraystretch{1.2}
        \centering
    \begin{tabularx}{0.95\linewidth}{*{1}{>{\arraybackslash}X}}
Context: Obrigado por entrar em contato com a Ajuda PRS-ORG, o meu nome é NAME-M. \\
Para começar, pode-me indicar o seu nome por favor? \\
Olá NAME-M! \\
Meu nome é NAME-M. \\
Hey NAME-M, nice meeting you. \\
Hope you are doing good today. \\
Sim estou lhe desejo que esteja bem também NAME-M.
Consegues ver os dos que enviei? \\ \\
   Translate the Brazilian Portuguese source text to English, given the context. \\
   Brazilian Portuguese: Anexos** \\
   English: \\ \\
  \TowerChat{} w/o Context: \textcolor{BadRed}{Anexos**} \\
  \TowerChat{} w/ Context: \textcolor{GoodGreen}{Attachments**} \\
   \bottomrule
    \end{tabularx}  
\caption{Full conversation of second example in Figure~\ref{fig:turn-analysis}. In \textcolor{BadRed}{red}, the translation without context, which is wrong; in \textcolor{GoodGreen}{green}, the translation with context, which is correct.} \label{fig:example_2}
\end{figure}

\begin{figure}[h]
         \small
         \renewcommand\tabularxcolumn[1]{m{#1}}% <-- added
        \renewcommand\arraystretch{1.2}
        \centering
    \begin{tabularx}{0.95\linewidth}{*{1}{>{\arraybackslash}X}}
Context: Bom dia
Obrigado por entrar em contato com a Ajuda PRS-ORG, o meu nome é NAME-M. \\
Para começar, pode-me indicar o seu nome por favor? \\
Meus PRS-ORGs points que comprei não caiu \\
Efetuei o pagamento e foi descontado \\
Mais ainda não caiu no jogo \\
Hello, nice to meet you. \\
I will surely try my best to help you with missing PRS-ORG Mobile points. \\ 
May I know when the purchase was made? \\
Então? \\
Hoje \\
Thanks for the details. \\
There's nothing to be worried about. \\ \\
Translate the Brazilian Portuguese source text to English, given the context. \\
   Brazilian Portuguese: Tou vendo \\
   English: \\ \\
  \TowerChat{} w/o Context: \textcolor{BadRed}{I'm watching.} \\
  \TowerChat{} w/ Context: \textcolor{GoodGreen}{I see} \\
   \bottomrule
    \end{tabularx}  
\caption{Full conversation where context helps improve the translation quality. In \textcolor{BadRed}{red}, the translation without context, which is wrong; in \textcolor{GoodGreen}{green}, the translation with context, which is correct.
In the translation without context, ``vendo'' is translated to ``watching''; this could be correct in certain contexts, but in this particular one---where the customer simply wants to acknowledge what the agent said---it is not. Instead, ``I see'' is the correct translation.} \label{fig:example_4}
\end{figure}

\begin{figure}[h]
         \small
         \renewcommand\tabularxcolumn[1]{m{#1}}% <-- added
        \renewcommand\arraystretch{1.2}
        \centering
    \begin{tabularx}{0.95\linewidth}{*{1}{>{\arraybackslash}X}}
Context: Obrigado por entrar em contato com a Ajuda PRS-ORG, o meu nome é NAME-M. \\
Para começar, pode-me indicar o seu nome por favor? \\
NAME-F \\
não consigo realizar o pagamento dos pacotes, dá não autorizado sendo que possuo crédito no cartão \\
Hello NAME-F, hope you are fine, how may I help you? \\
Hello NAME-F, hope you are fine. \\
I see, let me check if we can fix the issue that you are facing with making purchase. \\
Please share the email address linked to the account. \\
EMAIL \\
So what exactly happens when you go to make transaction? \\
passa o cartão, normalmente \\
segundos depois \\
a compra é estornada \\
tentei 2 cartões diferentes \\
Let me check. \\
On my end it comes as the transaction is pending. \\
o que devo fazer? \\
tentar depois? \\
Yes. \\ \\
Translate the English source text to Brazilian Portuguese, given the context. \\
   English: Give it 24 hours cool down time. \\
   Brazilian Portuguese: \\ \\
  \TowerChat{} w/o Context: \textcolor{BadRed}{Deixe por 24 horas para esfriar.} \\
  \TowerChat{} w/ Context: \textcolor{GoodGreen}{Dê um período de 24 horas de espera.} \\
   \bottomrule
    \end{tabularx}  
\caption{Full conversation where context helps improve the translation quality. In \textcolor{BadRed}{red}, the translation without context, which is wrong; in \textcolor{GoodGreen}{green}, the translation with context, which is correct.
In the translation without context, ``cool down'' is literally translated to ``esfriar'', which is incorrect in context---the agent is telling the customer to wait (``espera'').} \label{fig:example_3}
\end{figure}

\end{document}